\documentclass[default,iicol,sn-basic]{sn-jnl}

\usepackage{blindtext}

\usepackage{bm}

\renewcommand{\vec}[1]{\bm{\mathrm{#1}}}
\newcommand{\mat}[1]{\bm{\mathrm{#1}}}

\newcommand{\mathsc}[1]{\text{\normalfont\scshape#1}}

\newcommand{\Tau}{\mathcal{T}}
\newcommand{\blank}{\text{\#}}
\newcommand{\singlequote}[1]{`#1'}

\usepackage{caption}
\usepackage{subcaption}

\usepackage[frozencache,cachedir=minted-cache]{minted}

\jyear{2021}%

\theoremstyle{thmstyleone}%
\theoremstyle{thmstyletwo}%

\theoremstyle{thmstylethree}%

\begin{document}

\title[SoftCTC]{SoftCTC -- Semi-Supervised Learning for Text Recognition using Soft Pseudo-Labels}


\author*[*]{\fnm{Martin} \sur{Kišš} \sfx{\textsuperscript{[0000-0001-6853-0508]}}} \email{ikiss@fit.vut.cz}

\author[1]{\fnm{Michal} \sur{Hradiš} \sfx{\textsuperscript{[0000-0002-6364-129X]}}} \email{ihradis@fit.vut.cz}

\author[1]{\fnm{Karel} \sur{Beneš} \sfx{\textsuperscript{[0000-0002-0805-1860]}}} \email{ibenes@fit.vut.cz}

\author[1]{\fnm{Petr} \sur{Buchal}}

\author[1]{\fnm{Michal} \sur{Kula} \sfx{\textsuperscript{[0000-0002-5795-5938]}}} \email{ikula@fit.vut.cz}


\affil[1]{\orgdiv{Faculty of Information Technology}, \orgname{Brno University of Technology}, \orgaddress{\street{Božetěchova 1/2}, \city{Brno}, \postcode{612 00}, \country{Czech~Republic}}}


\abstract{
    This paper explores semi-supervised training for sequence tasks, such as Optical Character Recognition or Automatic Speech Recognition.
    We propose a novel loss function -- SoftCTC -- which is an extension of CTC allowing to consider multiple transcription variants at the same time.
    This allows to omit the confidence-based filtering step which is otherwise a crucial component of pseudo-labeling approaches to semi-supervised learning.
    We demonstrate the effectiveness of our method on a challenging handwriting recognition task and conclude that SoftCTC matches the performance of a finely-tuned filtering-based pipeline.
    We also evaluated SoftCTC in terms of computational efficiency, concluding that it is significantly more efficient than a naïve CTC-based approach for training on multiple transcription variants, and we make our GPU implementation public.
}




\maketitle

\section{Introduction}\label{sec:introduction}

Optical character recognition (OCR) is one of the fundamental tasks in computer vision.
Modern text recognition systems utilizing neural networks are usually based on one of two architectures -- a CRNN~\cite{shi_end--end_2017} architecture trained using a CTC~\cite{GravesCTC} loss function or a sequence-to-sequence~\cite{SutskeverSeq2Seq} architecture trained using an autoregressive factorization of the posterior probability of the ground-truth transcription.
To train such OCR system to a reasonable accuracy, a large amount of annotated data is required.

Recently, for other tasks than OCR, ap\-proach\-es utilizing huge datasets show great success in terms of precision and also the robustness of the trained system.
These approaches usually leverage vast amounts of data from the internet, either unlabeled or with annotations of uncertain quality.
Representatives of such approaches are ASR system Whisper~\cite{radford_robust_2022}, language model GPT-3~\cite{brown_language_2020}, or generative text-to-image models Stable Diffusion~\cite{rombach_high-resolution_2022} and DALLE-2~\cite{ramesh_hierarchical_2022}.
However, obtaining reasonably good quality transcriptions for training in a particular domain is still very time and/or money-consuming.

On the other hand, many data sources exist, from which a large number of scanned documents can be obtained, e.g. manuscripts of several authors from a certain period.
It would be useful to use such documents but, due to the lack of transcriptions, they cannot be used directly and a semi-supervised approach has to be taken.
In this paper, we propose a novel approach for adapting the OCR system to a specific target domain without the need for large-scale manual transcriptions.

We consider target and related domains in our semi-supervised scenario.
The target domain is the one where a large quantity of data is available, but only a small portion of it is annotated.
On the other hand, the related domain contains a sufficient amount of annotated data with similar characteristics, but models trained on it do not yield satisfying performance in terms of accuracy on the target domain.

As our baseline, we take a CRNN+CTC-based pseudo-labeling approach, such as studied by Kišš et al.~\cite{kiss_at-st_2021}.
There, an initialization model (also called a seed model) is first trained on the annotated data from both domains.
This model is then used to produce pseudo-labels for the unannotated part of the target domain, and, as the pseudo-labels are usually erroneous, only data with labels confident enough are added to the next training iterations with strong image augmentation.
Here a key role is played by the algorithm used to identify the labels confident enough.
Unfortunately, as the model might be very over-confident in its predictions~\cite{arazo_pseudo-labeling_2020}, selecting a suitable algorithm and tuning its hyper-parameters is not a trivial task.

In this paper, we propose to get rid of the confidence-based filtering step of the pseudo-la\-beling process.
Instead, for each sample in the unannotated target domain, we use prefix search decoding on the output probabilities provided by the seed model to produce an ensemble of transcription variants and we train the new model using all the variants at the same time.
This could be achieved by repeated computation of the CTC loss, but as we show in this paper, it is a very time-consuming method.
Instead, we propose a novel SoftCTC loss function that computes the loss of the model's output w.r.t. all transcription variants simultaneously.
The efficient formulation of the loss function allows it to consider astronomical numbers of transcription hypotheses\footnote{In our training setup, we consider on average ca. $10^{85}$ possible transcription variants for each text line; the respected \emph{Eddington number} estimates the number of protons in the observable universe to about $10^{80}$.}, overcoming the painful speed\,--\,accuracy trade-off of the naïve method.

Contributions of the paper are as follows:
\begin{enumerate}
    \item pseudo-labeling approach which does not rely on the confidence measure, on par with SotA
    \item a generalized version of the CTC loss -- SoftCTC -- which can handle multiple transcription variants simultaneously
    \item a formal description of encoding confusion networks for SoftCTC
    \item efficient GPU implementation of SoftCTC, publicly available at GitHub\footnote{\url{https://github.com/DCGM/SoftCTC}}.
\end{enumerate}

\section{Related Work}
\label{sec:related_work}

Semi-supervised learning uses both annotated and unannotated data for optimizing a model.
The unannotated data are usually used either for \emph{consistency regularization}~\cite{bachman_learning_2014, sajjadi_regularization_2016, tarvainen_mean_2017, berthelot_mixmatch_2019, kurakin_remixmatch_2020, xie_unsupervised_2020, englesson_generalized_2021, zheng_pushing_2022, aberdam_multimodal_2022}, \emph{pseudo-labeling}~\cite{lee_pseudo-label_2013, arazo_pseudo-labeling_2020, xie_self-training_2020, pham_meta_2021, nagai_recognizing_2020, stuner_self-training_2017, leifert_two_2020, das_adapting_2020, kiss_at-st_2021} (also called \emph{self-training}), or their combination~\cite{sohn_fixmatch_2020, weninger_semi-supervised_2020, wolf_self-training_2022}.
In general, all of these approaches train in a supervised fashion on the annotated data and differ in how the unlabeled data is utilized.

\paragraph*{Consistency regularization}
In the consistency regularization approach, given a training sample, the model is trained to produce the same output under different perturbations of the sample.
An important condition in this training is that the perturbations must preserve the semantics of the sample.
If this condition is met, the network is forced to focus on the useful content of the sample.

The idea of consistency regularization was first proposed by Bachman et al.~\cite{bachman_learning_2014} for training only on annotated data for image classification, and Sajjadi et al.~\cite{sajjadi_regularization_2016} redesigned the loss for training on unlabeled data.
In the Mean Teacher~\cite{tarvainen_mean_2017} approach, the consistency regularization is applied in the teacher-student setup where the student is trained on the labeled samples and at the same time, to produce the same output as the teacher for the unlabeled data.
The teacher is updated as an exponential moving average of student model weights.

In MixMatch~\cite{berthelot_mixmatch_2019} and ReMixMatch~\cite{kurakin_remixmatch_2020}, the consistency regularization is used for training of an image classifier on unannotated data; in both cases, the MixUp augmentation~\cite{zhang_mixup_2018} is used.
In MixMatch, the objective is to minimize the distance between the output produced for the unaugmented sample and the ``guessed'' label, which is a sharpened average of $K$ outputs produced for $K$ augmented versions of the same sample.
In the ReMixMatch approach, weak and strong augmentations are applied to the unlabeled sample and the distance between the two respective outputs is minimized.
Moreover, the output distribution for the weakly augmented sample is sharpened and adjusted according to the distribution of classes in the labeled dataset and the distribution of predictions on the unlabeled dataset.
An approach similar to ReMixMatch proposed by Xie et al. is called UDA~\cite{xie_unsupervised_2020} where the training on the unlabeled data is performed only if the confidence of the output for the weakly augmented version of the sample is higher than a predefined threshold.

Consistency regularization can also be used when training on noisy labels. 
Englesson et al.~\cite{englesson_generalized_2021} proposed a loss function based on the Jensen-Shannon divergence where the distance between two outputs and also the ground-truth labels is minimized.

\paragraph*{Pseudo-labeling}
The pseudo-labeling approach starts with an initialization model (sometimes also called a seed model) trained on annotated data using supervised learning.
This model is then used for predicting labels of each sample in an unannotated dataset (thus called \emph{pseudo-label}).
The sample together with its pseudo-label is then added to the annotated training dataset and further optimization of either a new model or continued training of the old model follows.
Optionally, a threshold, which controls whether the pseudo-label is confident enough to be used for future training, can be defined.
This whole process can also be seen as a teacher\,--\,student setup where the trained teacher produces labels for the unannotated data which are used together with the annotated data for training the student.

The idea of the pseudo-labels was proposed by Dong-Hyun Lee~\cite{lee_pseudo-label_2013} who used predicted hard pseudo-labels of unlabeled data for further optimization of a neural network.
In more recent work, Arazo et al.~\cite{arazo_pseudo-labeling_2020} proposes two regularizations and usage of the MixUp augmentation~\cite{zhang_mixup_2018} for suppressing the effect of confirmation bias.
As the result of the MixUp augmentation is a soft label (i.e. a relative class membership instead of a hard assignment), the difference to the previous approach lies in training on soft pseudo-labels.
The NoisyStudent~\cite{xie_self-training_2020} approach follows the teacher\,--\,student setup, where the student is equal to or larger than the teacher.
To train a more powerful student than the teacher, two types of noise are used during the training of the student: input noise (data augmentation) and model noise (dropout and stochastic depth).
Pham et al. proposed the Meta Pseudo Labels~\cite{pham_meta_2021} approach where both the teacher and the student are updated at the same time.
Firstly, the teacher predicts the pseudo-labels and then the student is trained using them.
If the student is trained using directly the soft pseudo-labels produced by the teacher, the standard back-propagation can be used to calculate the gradients of the teacher.
However, when the hard pseudo-labels are used, a reinforcement learning approach is used to calculate teacher's gradients.
\bigskip

Besides approaches utilizing either consistency regularization or pseudo-labeling, approaches combining both techniques exist as well.
Sohn et al. proposed the FixMatch~\cite{sohn_fixmatch_2020} approach in which weak and strong augmentations are applied to an unlabeled sample.
If the highest probability in the predicted distribution for the weakly augmented sample is above a certain threshold, the distribution is transformed into a hard label and used as a pseudo-label for the strongly augmented sample.
Adaptation of the FixMatch and the NoisyStudent approaches for Automatic Speech Recognition (ASR) was proposed by Weninger et al.~\cite{weninger_semi-supervised_2020}.
The difference between their approach and the FixMatch is that they use a pretrained teacher model for generating pseudo-labels and they also explore the usage of both the soft and the hard pseudo-labels.

\paragraph*{Semi-supervised learning in text recognition}
Both consistency regularization and pseudo-label\-ing were applied to text recognition.

Zheng et al.~\cite{zheng_pushing_2022} proposed an approach for semi-supervised learning of a scene text recognizer that uses consistency regularization in a similar fashion as the Mean Teacher approach.
Their architecture consists of two models (an online model and a target model) where the target model is an exponential moving average of the online model.
The online model is trained to recognize the annotated samples and to produce the same output under a strong augmentation as the target model under a weak augmentation for a given sample.
They also incorporate a domain adaptation loss into the training procedure.
Aberdam et al.~\cite{aberdam_multimodal_2022} proposed a multimodal semi-supervised learning for text recognition.
In their approach, they use a language model for correcting the text recognizer output.
During training, they use weak and strong augmentations to train both models simultaneously using consistency regularization.

Nagai~\cite{nagai_recognizing_2020} proposed a pseudo-labeling approach to recognizing historical Japanese cursive using an ensemble of $N$ recognizers trained on the same training data.
Then, the $N$ recognizers produce $N$ transcriptions for each unannotated sample and the final pseudo-label is produced by refinement implemented as a local search with a language model.
Finally, all recognizers are fine-tuned using the pseudo-labels.
A similar approach was proposed by Frinken et al.~\cite{frinken_evaluating_2009} where they studied confidence-based metrics together with a voting mechanism.
Three similar approaches for filtering out potentially erroneous transcriptions using a language model or a lexicon of the target language were proposed by Stuner et al.~\cite{stuner_self-training_2017}, Leifert et al.~\cite{leifert_two_2020} and Das et al.~\cite{das_adapting_2020}.
The first one uses a lexicon and the pseudo-label is kept for future training if the predicted transcription is found in the lexicon.
The second approach calculates the posterior probability of the predicted transcription at the character and the sentence level and if any of them is less than a certain threshold, the pseudo-label is rejected from the subsequent training.
In the last approach, the confidence measure is computed as a weighted sum of two terms: a sum of character log-probabilities from the output of the model and a logarithm of the joint probability of the character sequence calculated by the language model.
Constum et al.~\cite{constum_recognition_2022} evaluated different sizes of student networks in self-training scenario.
Additionally, they exploited the closed-domain nature of their data and applied WFST-encoded language models to constrain the OCR output.
In AT-ST~\cite{kiss_at-st_2021}, Kišš et al. proposed several confidence measures not requiring a language model and experimentally showed that the posterior probability of a predicted transcription works very well to eliminate erroneous pseudo-labels.

Wolf et al.~\cite{wolf_self-training_2022} proposed an approach similar to the FixMatch which is adapted to the text recognition problem.
The approach starts by training an initial model on synthesized word images.
This model is then used to produce pseudo-labels for the unannotated dataset.
Erroneous pseudo-labels are filtered out by a confidence measure based on the probability of individual characters in the transcription.
In the subsequent training, they use consistency regularization similar to the FixMatch approach.

Gao et al.~\cite{gao_semi-supervised_2021} proposed a semi-supervised learning approach to train a text recognizer utilizing also reinforcement learning techniques.
They use a hybrid loss function consisting of cross-entropy for training a seq2seq text recognizer and two reward-based loss functions.
The first loss function (edit reward) is used when an annotated sample is presented to the network and it is simply calculated as the negative character error rate of the predicted transcription w.r.t. the ground-truth transcription.
The second loss (embedding reward) is calculated as the cosine similarity between an embedding of the input image and embedding of the predicted transcription which is transformed into embeddings using a separate CNN and RNN respectively.

In our proposed approach, we follow the pseudo-labeling procedure.
Particularly, we extend the CTC objective to training on soft pseudo-labels.
This allows us to omit any explicit filtering based on confidence without performance degradation.

\subsection{CTC loss function}
\label{sec:ctc}

\begin{figure*}
    \centering
    \begin{subfigure}{0.32\linewidth}
        \centering
        \includegraphics{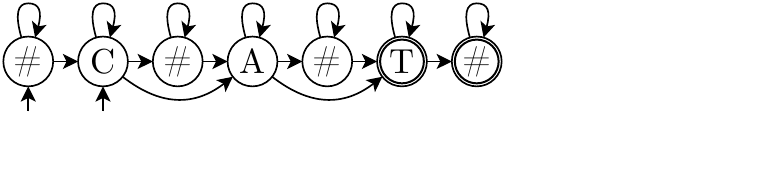}
        \caption{%
            Automaton representing labeling \singlequote{CAT}.
            Note that there are exactly two initial and two final states.
        }\label{fig:ctc-automaton}
    \end{subfigure}
    \hfill
    \begin{subfigure}{0.26\linewidth}
        \centering
        \includegraphics{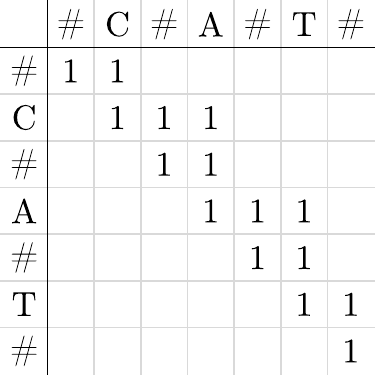}
        \caption{%
            The forward transition matrix $\vec{A}$.
        }\label{fig:ctc-cat-matrix}
    \end{subfigure}
    \hfill
    \begin{subfigure}{0.27\linewidth}
        \centering
        \includegraphics{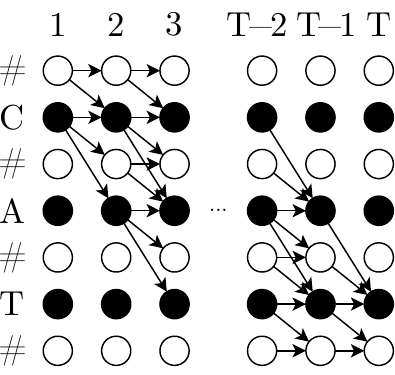}
        \caption{%
            Unrolled view of $\vec{\alpha}_t$ computation.
        }\label{fig:ctc-fwd-pass}
    \end{subfigure}
    \caption{%
        Different views of CTC illustrated for the labeling \singlequote{CAT}.
        Forward variables $\vec{\alpha}_t$ are calculated in the direction of the arrows, backward variables $\vec{\beta}_t$ are calculated in the opposite direction.
        The symbol \blank{} represents the \emph{blank} symbol.
    }\label{fig:ctc}
\end{figure*}

The Connectionist Temporal Classification (CTC)~\cite{GravesCTC} loss function is generally used when training a model on unsegmented sequence data (i.e. sequence data where the position of labels is not known), for example when training OCR~\cite{shi_end--end_2017, kiss_at-st_2021} or ASR~\cite{GravesCTC,Soltau-CTC-ASR-2017,WatanabeHybridCTC} neural networks.
Given input $\mathbf{x}$, the output of a neural network $\mathcal{N}$ is a matrix $\mathbf{y} = \mathcal{N}(\mathbf{x})$ of shape $\lvert{}V\rvert \times T$, where $V$ is the output vocabulary including the \emph{blank} symbol and $T$ is the length of the output which is proportional to the length of the network input.
An element $y_t(k)$ of this matrix is interpreted as the probability of observing label $k$ at step $t$.
The probability of any given path $\pi$ through the matrix $\mathbf{y}$ which represents a single text string alignment is then calculated as:
\begin{equation}
    \label{eq:prob_pi_given_x}
    p(\pi \vert \mathbf{x}) = \prod\limits_{t=1}^T y_t(\pi_t)
\end{equation}

To transform a path to a labeling (sequence of labels), a many-to-one map $\mathcal{B}$ merges repeating symbols in the path and then removes all blanks. 
The inverse mapping $\mathcal{B}^{-1}(\mathbf{l})$ then defines the set of all paths which, when transformed using $\mathcal{B}$, result in the given labeling $\mathbf{l}$.
The probability of a given labeling is then defined as the sum of probabilities of all paths corresponding to it:
\begin{equation}
    \label{eq:prob_l_given_x_prod}
    p(\mathbf{l} \vert \mathbf{x}) = \sum\limits_{\pi \in \mathcal{B}^{-1}(\mathbf{l})} p(\pi \vert \mathbf{x})
\end{equation}
As the number of all possible paths that can be decoded as the correct labeling grows exponentially with the length of the network output $T$, the sum in Eq.~\eqref{eq:prob_l_given_x_prod} is computed using dynamic programming.
Specifically, the forward-backward algorithm calculates forward and backward variables, in effect recursively calculating the sum over paths corresponding to prefixes and suffixes of the labeling.
Forward variables $\alpha_t(s)$ are interpreted as the total probability of all possible prefixes of a given symbol $s$ at step $t$.
Similarly, backward variables $\beta_t(s)$ represent the total probability of all possible suffixes of a given symbol $s$ at step $t$, much like in Baum-Welch training of Hidden Markov Models~\cite{Rabiner1989}.

To include blanks in the paths, a modified sequence of labels $\mathbf{l}^{\prime}$ is customarily constructed by interleaving the labels with \emph{blanks} and adding blanks also at the beginning and at the end of the sequence.
The length of this new sequence is $2\vert \mathbf{l} \vert + 1$ where $\mathbf{l}$ is the original labeling.
In all the following equations, symbol \singlequote{\blank} denotes the blank symbol.

Alternatively, the alignment $\pi$ can be seen in terms of finding a path through a finite state automaton (Fig.~\ref{fig:ctc-automaton}), where each state $i$ is associated with a symbol $X_i$.
The automaton contains both \emph{letter states} and \emph{blank states}.
Equivalently, the connections in such an automaton can be represented by a transition matrix $\vec{A}$ (Fig. \ref{fig:ctc-cat-matrix}).
Utilizing this notation, the forward variable $\vec{\alpha}_t$ can be defined:
\begin{equation}
    \label{eq:alphas}
    \vec{\alpha}_t = \vec{\alpha}_{t-1} \vec{A} \odot  \vec{q}_t
\end{equation}
Here, $\odot$ denotes element-wise product and $\vec{q}_t$ is the vector of state probabilities corresponding to the $\vec{y}_t$, i.e.\ $q_t(i)$ is the probability of symbol $X_i$ estimated by the neural network at step $t$.
Note that this is not a one-to-one mapping as not all symbols from $V$ need to occur in every text line and conversely, some may be repeated.
Most notably, the blank symbol corresponds to more than half of the states.

The recursion is initialized with:
\begin{align}\label{eq:alphas_init}
    \vec{\alpha}_1 &= \big[q_1(1), q_1(2), 0, 0, \ldots\big] \nonumber \\
                   &= \big[y_1(\blank), y_1(\mathbf{l}_1), 0, 0, \ldots\big]
\end{align}
This corresponds to allowing the first output frame to align to either the initial blank or the first letter of $\mathbf{l}$ with the corresponding probability from the respective entries in $y_1$.
The effective propagation of non-zero values of $\vec{\alpha}_t$ is illustrated in Fig.~\ref{fig:ctc-fwd-pass}.

Symmetrically, the matrix $\mat{A}^T$ describes the reverse of the accepting automaton.
Then the backward variable $\vec{\beta}_t$ can be defined as:
\begin{equation}
    \label{eq:betas}
    \vec{\beta}_t = \vec{\beta}_{t+1} \vec{A}^T \odot \vec{q}_t
\end{equation}
\begin{align}
    \label{eq:betas_init}
    \vec{\beta}_T &= \big[\ldots, 0, 0, q_T( \lvert\vec{l^\prime}\rvert-1), q_T(\lvert\vec{l^\prime}\rvert)\big] \nonumber\\
                  &= \big[\ldots, 0, 0, y_T(\mathbf{l}_{\vert \mathbf{l} \vert}), y_T(\blank)\big]
\end{align}

For any given $t$ and $s$, the product $\alpha_t(s) \beta_t(s)$ is proportional to the posterior probability that an alignment of $\vec{l'}$ to $\vec{x}$ goes through state $s$ in step $t$.
The factor of proportion is $q_t(s)$ because the corresponding output of the neural network is included in both $\alpha_t(s)$ and $\beta_t(s)$\footnote{%
In HMM literature, the issue is avoided by not including $\vec{q}_t$ in $\vec{\beta}_t$ -- at the cost of breaking the symmetry of the recursive formulations of $\vec{\alpha}_t$ and $\vec{\beta}_t$ and their initial values $\vec{\alpha}_1$ and $\vec{\beta}_T$~\cite{Rabiner1989}.
}.
Canceling $q_t(s)$ out once, we obtain the probability of the labeling $\vec{l}$ as the sum of these posterior probabilities over all states at any step $t$:
\begin{equation}
    \label{eq:prob_l_given_x_alpha_beta}
        \mathsc{CTC}(\vec{l}, \vec{x}) = p(\mathbf{l} \vert \mathbf{x}) = 
        \sum_{s=1}^{\vert \vec{l'}\vert}\frac{\alpha_t(s) \beta_t(s)}{q_t(s)}
\end{equation}

Effectively, CTC uses the posteriors implied by $\vec{\alpha} \odot \vec{\beta}$ to obtain soft targets for the frame-wise output of the neural network.
Since $\log p(\mathbf{l} \vert \mathbf{x})$ is typically optimized in practice, the objective function on the frame level is in fact cross-entropy.

\subsection{Models for capturing variants in transcriptions}

\begin{figure}[t]
    \begin{subfigure}{\linewidth}
        \centering
        \includegraphics[width=\linewidth]{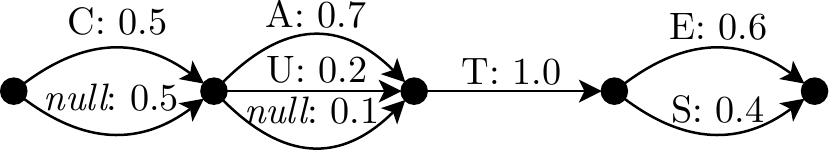}
        \caption{Confusion network}
        \label{fig:confusion-network}
    \end{subfigure}
    \\
    \begin{subfigure}{\linewidth}
        \centering
        \includegraphics[width=\linewidth]{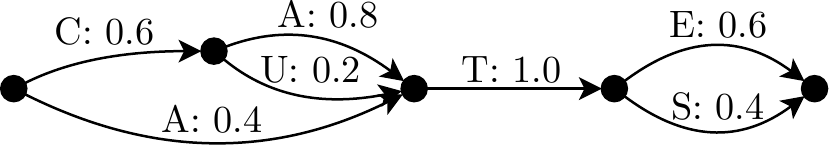}
        \caption{Lattice}
        \label{fig:lattice}
    \end{subfigure}
    \begin{subfigure}{\linewidth}
        \centering
        \includegraphics[width=\linewidth]{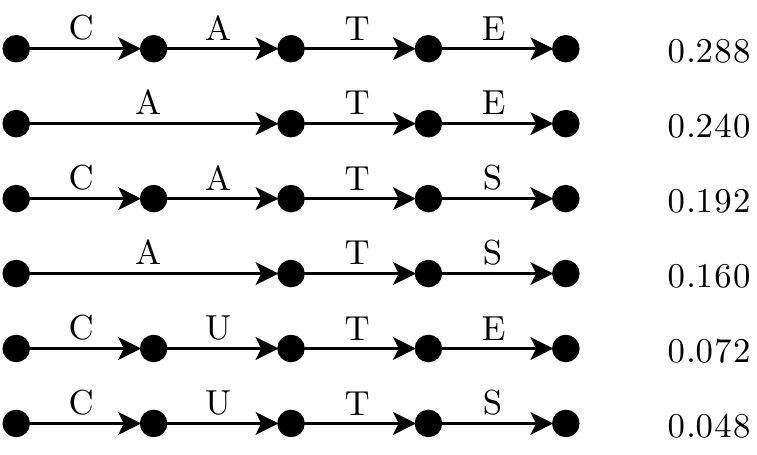}
        \caption{$n$-best list}
        \label{fig:n-best}
    \end{subfigure}
    \caption{%
        Confusion network and lattice. 
        The confusion network in (a) represents 12 different strings -- \singlequote{CATE}, \singlequote{ATE}, \singlequote{CATS}, \singlequote{CUTE}, etc. 
        However, it also represents \singlequote{UTE} which might not be desirable. 
        On the other hand, the lattice in (b) represents only 6 different strings (\singlequote{CATE}, \singlequote{CUTE}, \singlequote{CATS}, \singlequote{CUTS}, \singlequote{ATE}, and \singlequote{ATS}), but does not represent the string \singlequote{UTE} as the letter \singlequote{U} must be preceded by the letter \singlequote{C}.
        The $n$-best list in~(c) captures the same variants as the lattice, but they are treated as independent and they are only scored as a whole.
    }
    \label{fig:confusion-netwrok_lattice}
\end{figure}

Our proposed approach allows to train a text recognizer on multiple variants of a transcription.
The most straightforward representation of such is a collection of the variants along with their weights.
Assuming these are top variants according to some model, this is the so-called $n$-best list (Fig.~\ref{fig:n-best}).
It is usually readily available, but it is very wasteful in storage.

To capture transcription variants effectively, two common models exist -- \emph{confusion networks} (Fig.~\ref{fig:confusion-network}) and \emph{lattices} (Fig.~\ref{fig:lattice})~\cite{bangalore_computing_2001,rosti_combining_2007,fiscus_post-processing_1997,mangu_finding_2000}.
Both models can be represented as a directed acyclic multigraph where edges between two nodes contain local variants of the transcription called a \emph{confusion set}~\cite{mangu_finding_2000}.
In general, items in the confusion sets can be at any level of granularity, but usually it is at word, word-piece, or character level.

The main difference between confusion networks and lattices is that the latter captures sequential dependencies.
In confusion networks, all edges coming out of a node join immediately in the following node, while in lattices edges can end in different nodes.
Thus, transcriptions represented by a confusion network might contain sequences of characters which do not usually occur in a language.
In confusion networks, besides the items from the vocabulary, a confusion set can contain also a \emph{null} alternative with the meaning of an empty string (i.e. the confusion set can be skipped).
Note that this null alternative has no relation to the blank symbol from the CTC loss function.
In lattices, the null alternative is not present as it is possible to skip a character alternative via appropriate nodes and edges between them.
This results in the fact that lattices have stronger modeling power as they do not discard the sequential information and thus they can model the transcription variants more precisely than confusion networks.

In this work, our base recognition systems are based on CTC systems which do not explicitly model dependencies between output symbols.
Therefore, we focus on confusion networks rather than lattices.

\section{Utilizing Multiple Transcriptions in Training}
\label{sec:multi_ctc}

The CTC loss assumes that the model under training is optimized against a single ground-truth transcription (labeling $\vec{l}$).
This assumption is reflected directly in the formulas of forward and backward variables.
Still, in the scenario with multiple transcription variants, it is possible to apply CTC to training with multiple transcription variants:
One can compute the CTC loss w.r.t.\ all of the variants and then use their weighted sum\footnote{More precisely, as the loss is log-likelihood of the transcription given the input, it requires to exponentiate the loss, calculate the weighted sum, and then take the logarithm of the result. I.e., to perform the weighting in the probability domain.} as the objective function to be optimized.

Formally, let the target $L$ be a collection of labelings $\mathbf{l}_i$ with their associated probabilities $w_i$.
Then the objective function is defined as:
\begin{equation}\label{eq:multi-ctc}
    \mathsc{MultiCTC}(L, \vec{x}) = \sum_{i} w_i\, \mathsc{CTC}(\mathbf{l}_i, \vec{x}) 
\end{equation}

This approach corresponds directly to having $L$ represented as an $n$-best list.
Thus, it has two significant practical drawbacks:
(1) it requires a substantial increase in either computation time or memory, both proportional to the number of transcription variants in $L$, because
(2) it usually calculates the same value many times.
E.g.\ if two transcription variants differ only in the last character, the values of the forward variable will be the same until the very last character.
Backward variables will differ in values, but the majority of the computation is also shared, imagine the recursive Eq.~\eqref{eq:betas} unrolled and evaluated from left, using the associative property of matrix multiplication.

\subsection{SoftCTC}
To remedy this inefficiency, we propose SoftCTC loss function.
SoftCTC is a generalization of the CTC loss function for training a model on multiple variants of a transcription.
The main difference is that in SoftCTC, the accepting automaton can contain branching to accommodate for multiple, possibly local, variants of transcription.
Subsequently, the transition matrix $\mat{A}$ is no longer required to be binary with nonzero values limited to the main diagonal and two more above.
We still require that $\mat{A}$ is an upper triangular matrix so that the corresponding automaton is, apart from self-loops, acyclic\footnote{It is not a necessary condition, but any acyclic automaton can be expressed as a triangular matrix using topographic ordering and this way it is easy to check for possible cycles.}.
Further, we require that there is no increase in outgoing weights per state compared to a linear CTC automaton.
I.e.\ if there are multiple edges leading from a state to another, their weights need to sum up to one, with the exception of edges from a letter state skipping a blank, when the total outgoing value needs to equal to 3, as in vanilla CTC.

The second change is in definition of $\vec{\alpha}_1$ and $\vec{\beta}_T$.
Since the target $L$ can possibly allow for multiple starts and/or ends, there can be nonzero values in arbitrary positions, as defined by $L$.
The specific values are then given by entries in $\vec{y}_1$ and $\vec{y}_T$ corresponding to the states which are allowed as initial and/or final.

Once these conditions are met, the recursion in Eq.~\eqref{eq:alphas} and Eq.~\eqref{eq:betas} can be applied as is and the eventual objective function is:
\begin{equation}
    \mathsc{SoftCTC}(L, \vec{x}) = \sum_{s=1}^{\vert \vec{l'}\vert}\frac{\alpha_t(s) \beta_t(s)}{q_t(s)}
\end{equation}

Encoding the transcription variants directly into $\mat{A}$ allows efficient sharing of computation wherever applicable, thus breaking the linear relation between the number of considered variants, allowing to consider enormous numbers of variants in the training.

\section{Encoding transcription variants for SoftCTC}\label{sec:softctc}

In this section, we open by demonstrating how an $n$-best list can be encoded in the transition matrix $\mat{A}$.
Then, we show how the output of a regular CTC system can be efficiently represented in the form of a confusion network.
Finally, we describe how to construct the transition matrix from confusion networks.

\subsection{Encoding $n$-best lists}
\begin{figure}
    \centering
    \begin{subfigure}{1.0\linewidth}
        \centering
        \includegraphics{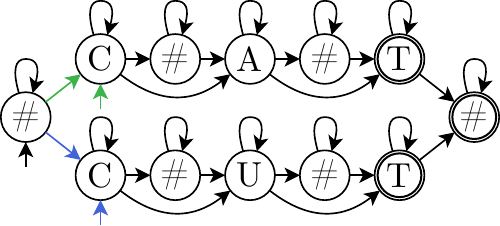}
        \caption{%
            Automaton representation of an $n$-best list.
            Note that the alternative initial letter states have weights corresponding to the probabilities of the variants.
        }
        \label{fig:softctc_nbest_automaton}
    \end{subfigure}
    \\
    \begin{subfigure}{1.0\linewidth}
        \centering
        \includegraphics{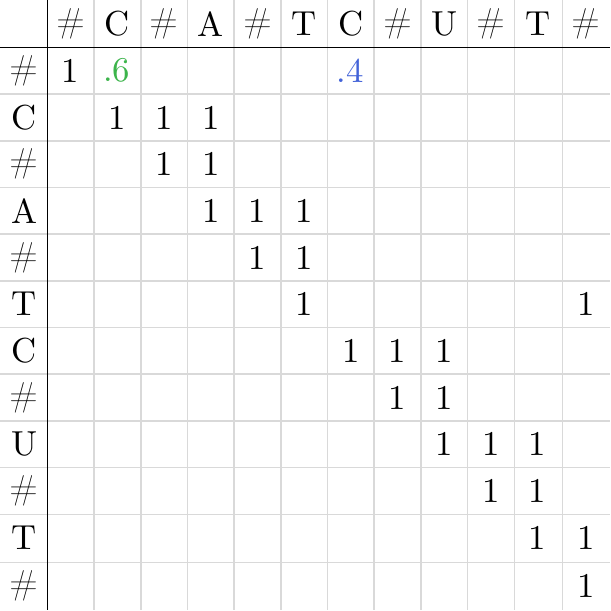}
        \hspace{0.1cm}
        \caption{Transition matrix representing the two separate branches in the automaton.}
        \label{fig:softctc_nbest_transition-matrix}
    \end{subfigure}

    \caption{%
        Encoding an $n$-best list containing \singlequote{CAT} and \singlequote{CUT} with weights 0.6 and 0.4 respectively.
    }
    \label{fig:softctc_nbests}
\end{figure}

The most straightforward variant representation to deal with is the $n$-best list.
It can be encoded as an automaton in a very straightforward way by collecting the variants in parallel and joining them by a single shared blank state\footnote{They can be even encoded as completely disjoint chains, but we prefer this way of presentation as it is easier to relate to encoding confusion networks.} at the beginning and at the end (Fig.~\ref{fig:softctc_nbest_automaton}).
Once the automaton is obtained, it can be directly encoded by a transition matrix (Fig.~\ref{fig:softctc_nbest_transition-matrix}) by distributing the variant probabilities to the edges leaving the initial blank state and following the regular CTC rules further on.
The values of $\vec{\alpha}_1$ need to reflect the possibility that a valid alignment $\pi$ begins directly in one of the letter states\,--\,with a corresponding probability\,--\,while the values of $\vec{\beta}_T$ must allow ending the alignment in any of the final letter states\,--\,without incurring any cost to such path.

Applying SoftCTC in this scenario is effectively the same as using the naive MultiCTC objective, with the additional cost of dealing with a huge, if sparse, matrix $\mat{A}$.
Also, it is trivial to see that SoftCTC is equivalent to CTC if there is a single transcription variant.

\subsection{Obtaining Confusion Networks}
\label{sec:softctc_obtaining_cns}

While $n$-best lists are highly inefficient as variant representations, they are readily available from prefix-search decoding of CTC output.
In this subsection, we show how to turn them into confusion networks.

The $n$-best list from prefix search decoding consists of a set of hypotheses along with their estimated\footnote{The prefix search does not take into account all possible paths corresponding to each transcription, so the actual posterior probabilities are higher. In practice, the amount of probability left out is negligible.} posterior probability.
We then construct the confusion network by iterating through the hypotheses in order of decreasing probability\footnote{Full implementation is public in the \texttt{pero-ocr} GitHub repository at \url{https://github.com/DCGM/pero-ocr}.}.
A pseudo-code of this algorithm is shown in Listing~\ref{algorithm:constructing-cn}.

\begin{listing}
\begin{minted}[fontsize=\small]{python}
def build_cn(nbests):
    nbests = sort_desc(nbests, key='score')    
    cn = build_linear_cn(nbests[0])

    for hyp in nbests[1:]:
        pivot = best_path(cn)
        alignment = align(pivot, hyp.text)
        add_path(cn, alignment, hyp.score)

    normalize_scores(cn)
    return cn
\end{minted}
\caption{Constructing a confusion network from an $n$-best list.}
\label{algorithm:constructing-cn}
\end{listing}

First, we take the most probable hypothesis and turn it into a trivial confusion network with no confusion yet.
Then, we repeatedly take the next hypothesis and compare it to the most probable path in the confusion network using Levenshtein alignment.
According to the obtained alignment, we add the hypothesis to the confusion network.
This can result in the introduction of new symbols into individual confusion sets or, in the case of insertion, in the construction of whole new confusion sets.
When adding a symbol to a confusion set, the hypothesis score is added to the current score of the corresponding alternative.
Finally, the scores are normalized in each confusion set.

\begin{figure}[t]
    \centering
    \includegraphics{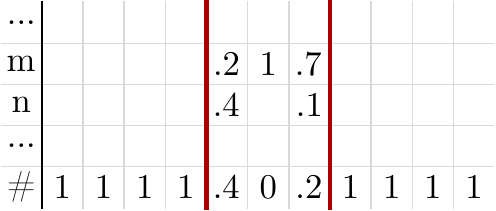}
    \caption{%
        Illustration of the partial line decoding strategy.
        The red lines delimit an unconfident sequence in the network's output where the prefix search decoding is used to obtain local variants.
    }
    \label{fig:partial_line_decoding_strategy}
\end{figure}

We propose two strategies to produce a confusion network for a single text line: \emph{full line strategy} and \emph{partial line strategy}.
In the former, the prefix search decoding is applied to the entire line.
In the latter, we only apply it to parts of the line that are not transcribed confidently enough while the rest of the line is decoded using greedy decoding.
In the network output $\vec{y}$, we define a confidently transcribed symbol as any symbol (including blank) which has a probability higher than 99\,\%.
Then, an unconfident part is defined as a continuous sequence in the network output between two confident blanks, which has at least one unconfident symbol, see Fig.~\ref{fig:partial_line_decoding_strategy}.
The confident parts of the output are decoded using greedy decoding and then they are transformed into trivial confusion networks (with only one variant between each two consecutive nodes).
Finally, confusion networks of all of the parts are concatenated into a single confusion network representing the whole line.

When generating a confusion network for a text line, the only parameter controlling the prefix search is the beam size, i.e.\ the number of partial hypotheses kept alive at any given moment of decoding.
Therefore, with a fixed amount of computing, the partial line strategy focuses on the low-confidence parts of the transcription more deeply because it is not wasting computation on carrying a larger beam through regions of high confidence, where only one path is really possible.

\begin{figure*}
    \centering
    \begin{subfigure}{1.0\textwidth}
        \centering
        \includegraphics[width=76mm]{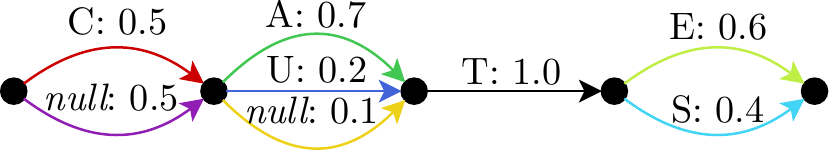}
        \caption{Confusion network}
        \label{fig:softctc_cn}
    \end{subfigure}
    \\
    \begin{subfigure}{1.0\textwidth}
        \centering
        \includegraphics[width=1.0\textwidth]{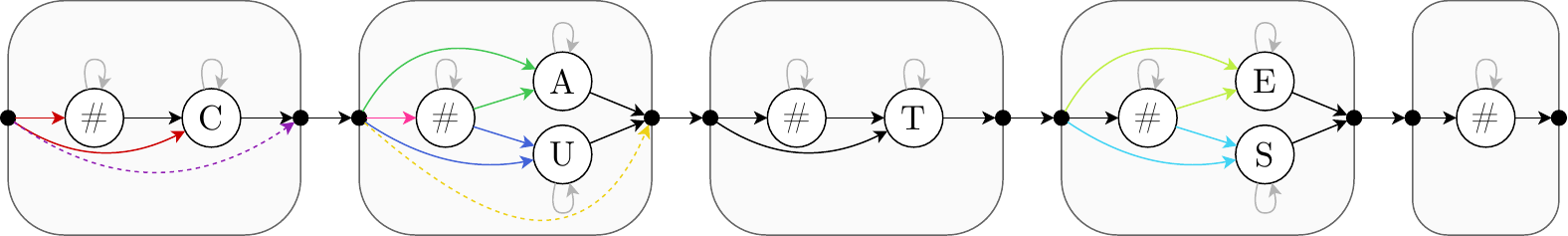}
        \caption{%
            Transcription confusion model.
            Black arrows, as well as the gray ones (self-loops), have the value of $1$ while the colored ones have the value from the confusion network above.
            In confusion character groups, the dashed line represents the $\epsilon$-transition.
        }
        \label{fig:softctc_model}
    \end{subfigure}
    \\
    \begin{subfigure}{0.66\textwidth}
        \centering
        \includegraphics{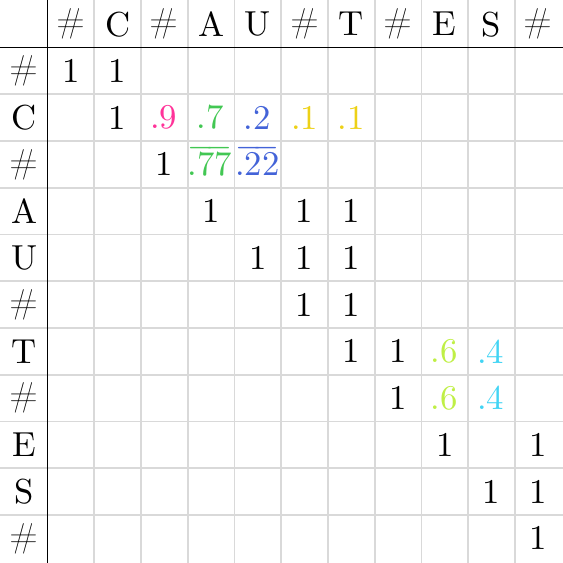}
        \hspace{0.1cm}
        \includegraphics{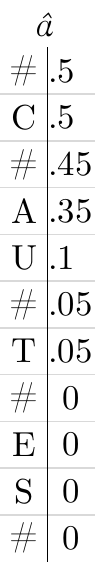}
        \hspace{0.1cm}
        \includegraphics{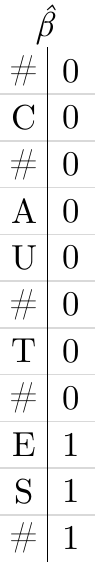}
        \caption{%
            Transition matrix $\mat{A}$ and vectors $\vec{\hat{\alpha}}$ and $\vec{\hat{\beta}}$ (Eq.~\ref{eq:alpha_hat} and~\ref{eq:beta_hat}).
        }
        \label{fig:softctc_transition-matrices}
    \end{subfigure}
    \hfill
    \begin{subfigure}{0.33\textwidth}
        \centering
        \includegraphics{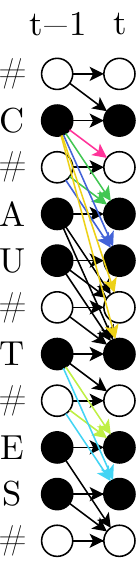}
        \caption{%
            Propagating $\vec{\alpha}_t$
        }
        \label{fig:softctc_structure}
    \end{subfigure}
    \caption{SoftCTC overview. 
    A confusion network (a) is transformed into transition matrices and initialization vectors (c) using a transcription confusion model (b).
    The forward-backward algorithm then uses the transition matrices and initialization vectors to calculate the forward and backward variables (d).
    }
    \label{fig:softctc_overview}
\end{figure*}

\subsection{Encoding Confusion Networks}
Despite its graphical nature, a confusion network (CN) differs from a CTC automaton in two aspects:
(1) All information in a CN is encoded on the edges, while a CTC automaton has symbols associated with states.
(2) Individual confusion sets in a CN may contain a \emph{null} edge allowing, possibly chained, skips of symbols, while each edge in a CTC automaton implies aligning one frame from $\vec{y}$.
To overcome these differences, we propose a two-step process of encoding a CN into a transition matrix $\mat{A}$\footnote{The whole process is implemented in the SoftCTC GitHub repository, specifically in  \url{https://github.com/DCGM/SoftCTC/blob/main/soft_ctc/models/connections.py}.}.

In the first step, a \emph{transcription confusion model} (TCM) is constructed, turning confusion sets into \emph{character confusion groups} which contain properly setup symbol- and blank-states and have a special treatment for the null edges in the input CN.
Then, we construct the transition matrix from the TCM.
Finally, the values of $\vec{\alpha}_1$ and $\vec{\beta}_T$ are also derived from the TCM.

\subsubsection*{Transcription confusion model}

A character confusion group (CCG) is a fragment of a CTC automaton which allows to align a slice of the output matrix $\mathbf{y}$ to the corresponding confusion set (Fig.~\ref{fig:softctc_model}).
In a direct representation of the letters in the confusion set, a CCG contains a blank state and a set of letter states, one for each letter in the confusion set.

However, the purpose of a CCG is not only to represent the alternative letters in a confusion set but also the possibility of this CCG being \emph{null}.
To this end, we introduce the notion of an \emph{entry point} and an \emph{exit point} of a CCG and we create a direct edge connecting the two -- the \emph{$\epsilon$-transition}.
The role of $\epsilon$-transition is the following:
If the CCG aligns to a letter, the path corresponding to it goes from the entry -- possibly through the blank state -- through the state of said letter to the exit point of the CCG.
If the CCG does not align to a letter, the $\epsilon$-transition is taken, bypassing all the actual states.
Note that entry and exit points are not actual states and will thus not be represented in the transition matrix $\mat{A}$.

The weight of a particular $\epsilon$-tran\-si\-tion $\epsilon_\tau$ is taken directly from the confusion network:
\begin{equation}
    p_{\tau}(\epsilon) = \mathit{CN}_{\tau}(null)
\end{equation}
where $\mathit{CN}_{\tau}(null)$ represents the probability of $null$ alternative in $\tau$-th confusion set of the confusion network $\mathit{CN}$.

With $p_{\tau}(\epsilon)$ defined, we proceed to define the probability of all symbols $X$ in the $\tau$-th CCG:
\begin{equation}
    p_{\tau}(X) = 
    \begin{cases}
        \mathit{CN}_\tau(X)        & X \neq \blank \\
        1 - p_{\tau}(\epsilon) & X = \blank
    \end{cases}
\end{equation}
where \blank{} denotes blank symbol.
Note that by the second case, we effectively prevent a null CCG to align to any blanks.

Finally, one extra CCG is appended to a TCM.
This CCG consists only of the blank state and it allows arbitrary blank predictions at the end of $\mathbf{y}$, again copying vanilla CTC behaviour.
The total number of all CCGs in a TCM is denoted as $\Tau$.

\subsubsection*{Transition matrix}\label{sec:softctc_fwd_bwd}
A transcription confusion model contains all the states which will constitute the SoftCTC target $L$.
It remains to properly encode the transition weights between them into the matrix $\mat{A}$ (Fig.~\ref{fig:softctc_transition-matrices}).

To keep $\mat{A}$ triangular, we order the states according to the left-to-right ordering of CCGs, with the blank state preceding letter states in each CCG.

Then, we define the entry and exit weight for each state in each CCG:
\begin{align}
\begin{split}
    \label{eq:p_in_forward}
    p_{\tau}^{\mathit{in}}(X) = {}& p_{\tau}(X)
\end{split} \\
    \label{eq:p_out_forward}
    p_{\tau}^{\mathit{out}}(X) = {}&
        \begin{cases}
        1 & X \neq \blank \\
        0 & X = \blank
        \end{cases}
\end{align}
where $p_{\tau}^{in}(X)$ represents the weight of an edge from the entry point to the symbol $X$ in $\tau$-th CCG and $p_{\tau}^{out}(X)$ represents the weight of an edge from the symbol $X$ to the exit point.
Note that by abuse of notation, we label all of these as $p(\cdot)$, but they are rather \emph{weights} than \emph{probabilities}.

Finally, we proceed to define transition weights between actual states.
Let $i$ and $j$ be two states in the TCM, originating from CCGs $\tau_i$ and $\tau_j$, associated with symbols $X_i$ and $X_j$.
Then, the transition matrix $\mat{A}$ is composed of weights $a_{i,j}$ defined as follows:
\begin{equation}
    \label{eq:p_X-Y}
    \begin{array}{l}
        a_{i,j} = \vspace{0.2cm}\\
        
        \hspace{0.2cm}
        \begin{cases}
            p_{\tau_i}^{\mathit{out}}(X_{i}) \cdot \prod\limits_{\tau=\tau_i + 1}^{\tau_j - 1} p_{\tau}(\epsilon) \cdot p_{\tau_j}^{\mathit{in}}(X_{j}) \\
                \hfill \tau_i < \tau_j, X_i \neq X_j \vspace{0.2cm}\\
            \frac{p_{\tau_j}^{\mathit{in}}(X_j)}{1-p_{\tau_j}(\epsilon)}
                \hfill \hspace{1.84cm} \tau_i = \tau_j, X_i = \blank, X_j \neq \blank  \\
            1   \hfill i = j \\
            0   \hfill \text{otherwise}
        \end{cases}
    \end{array}
\end{equation}
The interpretation of individual cases is as follows:
(1) A jump between CCGs.
Note that for $\tau_i + 1 = \tau_j$, the product in the middle is empty as we are not taking any $\epsilon$-transition.
On the other hand, it is possible that several $\epsilon$-transitions are chained and the alignment skips multiple letter states\footnote{This is very similar to $\epsilon$-closures in determinization of finite state automata.}.
(2) A movement from a blank state into a letter state within a single CCG.
(3) A self-loop.
Also observe that, like in CTC, it is not possible to bypass character states -- except when $\epsilon$-transition is taken.

\subsubsection*{Initial vectors}
Values of the initial forward vector $\vec{\alpha}_1$ are based on the possibility of the alignment to start in a particular state.
Symmetrically, the initial backward vector $\vec{\beta}_T$ is based on the possibility of the alignment ending in a particular state.

Given a state $i$ with symbol $X_i$ in the $\tau$-th CCG of a TCM, the values $\alpha_1(i)$ and $\beta_T(i)$ we define:
\begin{equation}\label{eq:alpha_hat}
    \hat{\alpha}(i) = \prod\limits_{\tau'=1}^{\tau-1} p_{\tau'}(\epsilon) \cdot p_{\tau}^{\mathit{in}}(X_i)
\end{equation}
\begin{equation}\label{eq:beta_hat}
    \hat{\beta}(i) = p_{\tau}^{\mathit{out}}(X_i) \cdot \prod\limits_{\tau'=\tau+1}^{\Tau-1} p_{\tau'}(\epsilon)
\end{equation}
Note that these are subterms of the first case in Eq.~\eqref{eq:p_X-Y}.
Also note, that $p_{\tau}^{\mathit{out}}(X_i)$ is a binary indicator (Eq.~\ref{eq:p_out_forward}), so $\beta_T$ only collects actual transition weights from the $\epsilon$-transitions.

Then the initial vectors are:
\begin{equation}\label{eq:softctc_init_alpha}
    \vec{\alpha}_1 = \vec{\hat{\alpha}} \odot \vec{q}_1
\end{equation}
\begin{equation}\label{eq:softctc_init_beta}
    \vec{\beta}_T = \vec{\hat{\beta}} \odot \vec{q}_T
\end{equation}
Comparing this to the initial vectors in vanilla CTC, the implied $\vec{\hat{\alpha}}$ and $\vec{\hat{\beta}}$ in Eq.~\ref{eq:alphas_init} and~\ref{eq:betas_init} are binary vectors describing the two initial and final states.

\subsection{Lattices}\label{sec:softctc_beyond_cn}
For lattices, one does not need to deal with $\epsilon$-transitions, because only actual symbols\footnote{The exception being a lattice representing that the whole line is possibly empty. This can however be easily addressed in a post-hoc fashion by constructing a direct connection from the first to the last blank state.} are captured in lattices.
Therefore, a lattice can be directly transformed into SoftCTC complete labeling:
Every edge in the lattice is represented by a connected pair of a blank state and a letter state.
The probability of the symbol given in the lattice is then encoded as the weight of edges leading to both of these states.
Links between these pairs of states always originate from letter states (to prevent skipping them) and follow the structure given by the states in the original lattice.

We do not experiment with lattices in this work as they are not a natural output of a CTC-based OCR system.

\section{Experiments}
\label{sec:experiments}

We experimentally demonstrate the effective\-ness of the proposed SoftCTC loss on a difficult hand\-writing recognition task.
In the conducted experiments, we follow the pseudo-labeling (self-training) process:
(1) we train a seed model on the mixture of annotated lines from a related domain dataset (READ'17) and a given target domain dataset (Bentham-Full, Bentham-20k, Rodrigo, READ'16),
(2) we use this model to produce confusion networks (i.e. soft pseudo-labels) for lines in the unannotated target domain dataset, and 
(3) we either train a new model or continue training on all data using the proposed SoftCTC loss function.

\begin{figure}
    \begin{subfigure}{\linewidth}
        \centering
        \includegraphics[width=\linewidth]{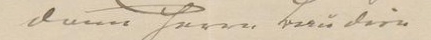}
        \caption{READ'17 dataset (related domain)}
        \label{fig:line-examples_read17}
    \end{subfigure}
    \\
    \begin{subfigure}{\linewidth}
        \centering
        \includegraphics[width=\linewidth]{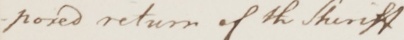}
        \caption{Bentham collection (target domain)}
        \label{fig:line-examples_bentham-unannotated}
    \end{subfigure}
    \\
    \begin{subfigure}{\linewidth}
        \centering
        \includegraphics[width=\linewidth]{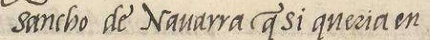}
        \caption{Rodrigo dataset (target domain)}
        \label{fig:line-examples_rodrigo}
    \end{subfigure}
    \\
    \begin{subfigure}{\linewidth}
        \centering
        \includegraphics[width=\linewidth]{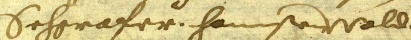}
        \caption{READ'16 dataset (target domain)}
        \label{fig:line-examples_read16}
    \end{subfigure}
    \caption{Examples of text lines from datasets.}
    \label{fig:dataset}
\end{figure}

For the first group of conducted experiments, where we study the properties and robustness of the proposed SoftCTC loss, we adopted datasets from a recent paper on self-training of OCR systems~\cite{kiss_at-st_2021}. 
Specifically, we use the \emph{handwritten--small setup}, the most challenging one, for all of the experiments.
In this setup, the related domain is the ICDAR 2017 READ Dataset~\cite{sanchez_icdar2017_2017} (READ'17) and for the target domain the annotated and unannotated datasets are the ICFHR 2014 Bentham Dataset~\cite{sanchez_icfhr2014_2014} and pages obtained from the Bentham Project\footnote{\url{https://www.ucl.ac.uk/bentham-project}}, respectively.
We call this entire dataset as Bentham-Full.
The READ'17 Dataset contains pages written in German, Italian, and French, while the pages in the Bentham-Full dataset are written in early 19\textsuperscript{th}-century English.
Table~\ref{tab:dataset} summarizes the dataset sizes and Fig.~\ref{fig:dataset} shows samples from the dataset.
Note that the number of training samples of the annotated target domain dataset in the table differs from the number presented in~\cite{kiss_at-st_2021} as we use just the small setup where only 10\,\% of training pages are used.

In the second group of experiments, we explore the characteristics and scaling abilities of the SoftCTC in a real use-case scenario.
In these experiments, we use the READ'17 as the related domain as well.
The target domain datasets are the Bentham-20k (a subset of the Bentham-Full dataset), Rodrigo~\cite{serrano_rodrigo_2010}, and ICFHR 2016 READ dataset~\cite{sanchez_icfhr2016_2016}.
The Rodrigo dataset comprises pages written by a single author in mid- 16\textsuperscript{th} century Spanish.
The READ'16 dataset comprises pages from the late 15\textsuperscript{th} to early 19\textsuperscript{th} century written in German containing minutes of council meetings written by an unknown number of writers.
As neither the Rodrigo database nor the READ'16 dataset initially contains unsupervised data, we use only a subset of the training lines (128, 256, 512, 1024, or 2048 lines) and we treat the rest of the training lines as the unsupervised ones.
As the READ'16 dataset and the original ICFHR 2014 Bentham Dataset are similar in the number of training lines, we decided to use additional 13k lines from the unannotated Bentham-Full collection, so we can show the results on differently sized datasets.
In total, the dataset contains about 20k unsupervised text lines and therefore we call this dataset Bentham-20k.

\begin{table}[t]
    \centering
    \caption{%
        Sizes of the used datasets in number of lines.
        For the training sets, we report the number of text lines used in our scenario, not the actual sizes in the original datasets.
    }\label{tab:dataset}
    \begin{tabular}{
        @{\extracolsep{2pt}}@{\kern\tabcolsep}
        lrcr}
        \toprule    

        Dataset
        & \multicolumn{1}{c}{Training}
        & \multicolumn{1}{c}{Validation}
        & \multicolumn{1}{c}{Test}
        \\
        
        \midrule
        READ'17 &    189\,805 & -- & 700 \\
        \addlinespace[0.1cm]
        Bentham-Full \\
        -- annotated   &         894 & 1\,415 & 860 \\
        -- unannotated & 1\,141\,566 & --     & \multicolumn{1}{c}{--}  \\
        \addlinespace[0.1cm]
        Bentham-20k \\
        -- annotated   & max 2\,048 & 1\,415 & 860 \\
        -- unannotated &    20\,146 & --     & \multicolumn{1}{c}{--}  \\
        \addlinespace[0.1cm]
        Rodrigo \\
        -- annotated   & max 2\,048 & 1\,000 & 5\,010 \\
        -- unannotated &    15\,546 & --     & \multicolumn{1}{c}{--}  \\
        \addlinespace[0.1cm]
        READ'16 \\
        -- annotated   & max 2\,048 & 1\,040 & 1\,138 \\
        -- unannotated &    6\,316 & --     & \multicolumn{1}{c}{--}  \\
        \bottomrule
    \end{tabular}
\end{table}

The rest of this section is organized as follows:
Section~\ref{sec:experiments_architecture} introduces the architecture of the neural network used in all conducted experiments.
In Section~\ref{sec:experiments_confusion_networks} we present details about generating a confusion network from the output of the neural network.
Sections~\ref{sec:experiments_group1} and~\ref{sec:experiments_group2} present the results of the conducted self-training experiments.
The last section compares the proposed SoftCTC loss with the MultiCTC loss in terms of computation efficiency.

\subsection{Optical model specification}
\label{sec:experiments_architecture}

The optical model is a CRNN-based neural network~\cite{shi_end--end_2017} optimized using the proposed SoftCTC loss or standard CTC loss~\cite{GravesCTC} for comparison.
The convolutional part consists of four convolutional blocks, each containing two 2D convolutional layers with ReLU activations followed by max-pooling and dropout layers.
To speed up the training, we use pretrained weights for the convolutional part from VGG16\footnote{From PyTorch module \texttt{torchvision.models.vgg16}.}.
The recurrent part is composed of three bidirectional LSTM layers where each processes the output of the convolutional part in increasingly smaller width resolution.
The output of each LSTM is then upsampled by nearest-neighbor interpolation back to the original resolution, they are summed up and finally processed by another bidirectional LSTM layer.
In all experiments, the input of the network is a text line image of shape $W \times 40$, where $W$ is the width of the image and its height is fixed to 40 pixels.
The output of the network is a matrix of shape $W/4 \times \lvert{}V\rvert$ where $V$ is the output vocabulary including the blank symbol.

\subsection{Generating confusion networks}
\label{sec:experiments_confusion_networks}

We generate the confusion networks as proposed in Section~\ref{sec:softctc_obtaining_cns}.
Orthogonal to the full and partial line strategy, we also experimented with constructing the confusion network from multiple variants of the same text line, each with a different augmentation applied.
We aimed to create even more variability in the transcriptions.
Using either of the strategies above, we converted each augmented version into a single confusion network, and then we merged all of the confusion networks.
The merging step was done similarly to the algorithm used when producing confusion networks.
The most probable transcriptions of both networks were aligned using Levenshtein alignment and the corresponding confusion sets were merged together.
After merging all the variants of transcription into a single confusion network, we normalized the edge scores.
This normalization step was not performed right after producing the individual confusion network but at only the very end to pronounce the overall transcription confidence of each augmented version in the final confusion network.

\subsection{Variability in soft-labels}
\label{sec:experiments_group1}

In the first group of experiments, we focus on the properties and robustness of the SoftCTC.
The first two parts describe experiments with the two decoding strategies proposed in~\ref{sec:softctc_obtaining_cns} and the experiments where the variability in transcriptions is increased using training augmentations during transcribing.
Next, we investigate the possibility to filter text line outliers (poor text lines) using the associated confusion networks.
Finally, we compare SoftCTC and MultiCTC with results from AT-ST\footnote{To compare the results as fairly as possible, we took the seed model directly from the AT-ST~\cite{kiss_at-st_2021}.},

As both proposed text line decoding strategies might potentially generate confusion networks with really low-probability character alternatives in their confusion sets, in all experiments we decided to keep only alternatives with probability higher than 1\,\%.

In each experiment, the optical model was optimized using Adam with an initial learning rate $2 \times 10^{-4}$.
The model was trained for 250k iterations and the learning rate was halved after 150k and 200k iterations.
During training, input images were augmented by blurring, adding noise, changing colors, applying affine transformations, and masking.
This exactly follows the training process described in~\cite{kiss_at-st_2021}.

\subsubsection*{Fighting the confirmation bias}
From prior research~\cite{arazo_pseudo-labeling_2020} and experience with end-to-end OCR models, we know that these models tend to suffer from over-confidence, manifested by overly sharp outputs.
Therefore we investigate the possibility of smoothing out the output probabilities in the first experiment.

We modify the individual probabilities in confusion networks by taking their $n$-th root and renormalizing them.
Equivalent to increasing the softmax temperature, increasing $n$ smooths the probabilities out.
Eventually, for $n=\infty$, all character alternatives in the particular confusion set are equally likely.
Note that this increases the role of the 1\,\% cut-off.

We also experimented with the proposed strategies for producing transcription hypotheses -- the \emph{full line} and \emph{partial line} strategies described in the previous section.
We chose the beam sizes to be 128 for the full line strategy and 16 for the partial line strategy. 
While it is difficult to compare these directly, this setting results in pretty much the same size of annotations.

Despite the smaller beam size value for the partial line strategy, the fact that the decoding process is possibly invoked multiple times per text line leads to higher overall diversity.
By combining the individual hypotheses into a confusion network, the total number of variants represented is increased even further:
On average, the full line strategy produces $1.29 \cdot 10^{17}$ variants per line.
For partial line strategy, the average is even higher at $2.18 \cdot 10^{85}$.

The results of these experiments are in Table~\ref{tab:results_adjustments}.
The results show small consistent improvements when used with a beam of size 16 and the partial decoding strategy.
However, the differences are rather negligible.

\begin{table}[t]
    \centering
    \caption{Results of experiments with the proposed decoding strategies and adjusting probabilities of character alternatives in confusion networks.
    The results are reported as CER [\%] on the validation and test sets of the Bentham-Full dataset.
    The first column contains the degree of the root used for smoothing the probabilities.}
    \label{tab:results_adjustments}
    \begin{tabular}{
        @{\extracolsep{4pt}}@{\kern\tabcolsep}
        ccccc}
        \toprule
        & \multicolumn{2}{c}{Beam 128 (full)} & \multicolumn{2}{c}{Beam 16 (partial)} \\
            \cline{2-3}
            \cline{4-5}
        n        & Val  & Test & Val  & Test \\
        \midrule
        1        & 4.59 & 4.52 & 4.62 & 4.58 \\
        2        & 4.62 & \textbf{4.35} & 4.60 & 4.46 \\
        3        & 4.65 & 4.46 & 4.64 & 4.49 \\
        4        & 4.62 & 4.43 & 4.64 & 4.43 \\
        5        & 4.67 & 4.46 & 4.55 & 4.41 \\
        8        & \textbf{4.57} & 4.44 & \textbf{4.53} & 4.37 \\
        $\infty$ & 4.67 & 4.45 & \textbf{4.53} & \textbf{4.30} \\
        \bottomrule
    \end{tabular}
\end{table}

\subsubsection*{Utilizing augmented text lines}
The aim of the second set of experiments is in increasing the number of transcription variants and investigating the impact on the resulting model.
The main idea lies in the second step of the pseudo-labeling process (generating pseudo-labels) where we use training augmentations during the transcribing phase.
In the experiments, we compare two augmentation settings -- with and without the masking augmentations proposed in AT-ST~\cite{kiss_at-st_2021}.
Without the masking augmentation, the seed model has the entire visual information in its input.
On the other hand, when the masking is present during the transcribing phase, it should create more variability in transcriptions as the seed model needs to infer the transcription of the missing parts from their surroundings.
As the transcription of the augmented lines depends on the applied transformations, we transcribe each line several times, we decode the output using the partial line strategy with beam size 16, and then we merge the individual confusion networks using the algorithm described in the previous section.

The results of these experiments are in Table~\ref{tab:results_merging}.
The results show that in these experiments the masking augmentation is not beneficial as it is strictly worse than the augmentation without masking.
Moreover, the results are in each case worse than the best results from the previous set of experiments.
We believe that the increased variability in transcriptions confuses the new model during training.

\begin{table}[t]
    \centering
    \caption{Results of experiments with generating confusion networks with applied augmentation. 
    The results are reported as CER [\%] on the validation and test sets of the Bentham-Full dataset.
    The number of merged confusion networks for a text line is in the first column. 
    }
    \label{tab:results_merging}
    \begin{tabular}{
        @{\extracolsep{4pt}}@{\kern\tabcolsep}
        rcccc}
        \toprule
        & \multicolumn{2}{c}{Without masking} & \multicolumn{2}{c}{With masking} \\
            \cline{2-3}
            \cline{4-5}
        \multicolumn{1}{c}{m} & Val & Test & Val & Test \\
        \midrule
        2 & \textbf{4.70} & \textbf{4.58} & \textbf{4.98} & \textbf{4.85} \\
        4 & 4.94 & 4.79 & 5.42 & 5.31 \\
        8 & 5.59 & 5.60 & 6.84 & 6.74 \\
        \bottomrule
    \end{tabular}
\end{table}

\subsubsection*{Filtering low-confidence lines}
The third set of experiments investigates the possibility of filtering line outliers (e.g. wrongly detected lines, poorly readable or unreadable lines, etc.) based on the produced confusion networks.
We define the metric to sort confusion networks as the product of sizes of confusion sets in a given confusion network $\mathit{CN}$ divided by the number of confusion sets $\Tau$:

\begin{equation}
    M(\mathit{CN}) = \frac{1}{\Tau}  \prod\limits_\tau \lvert \mathit{CN}_\tau \rvert
\end{equation}
The idea behind this metric is that the more transcription variants the confusion network represents, the more likely it is that the text line is an outlier.

We decode the lines of the unannotated dataset using a beam of size 16 and the partial decoding strategy and we sort them in ascending order based on its confusion networks.
Then we filter out the last 5\,\%, 10\,\%, and 20\,\% of the lines, respectively.
Finally, for each metric and each portion of remaining data we train a model with $n=\infty$.
The results in Table~\ref{tab:results_filtering} show that the lines filtering improves the model rather negligibly and only on the validation set.

\begin{table}[t]
    \centering
    \caption{Results of filtering poor lines, reported as CER [\%] on the validation and test sets of the Bentham-Full dataset.
    The first column represents the relative amount of text lines filtered out.
}
    \label{tab:results_filtering}
    \begin{tabular}{
        @{\extracolsep{4pt}}@{\kern\tabcolsep}
        rcc}
        \toprule
        \multicolumn{1}{c}{f} & Val & Test \\
        \midrule
         0\,\% & 4.53 & \textbf{4.30} \\
         5\,\% & 4.59 & 4.38 \\
        10\,\% & \textbf{4.48} & 4.35 \\
        20\,\% & 4.55 & 4.34 \\
        \bottomrule
    \end{tabular}
\end{table}

\subsubsection*{Comparison with AT-ST and MultiCTC}
Finally, we compare the best results of the SoftCTC with the AT-ST~\cite{kiss_at-st_2021} approach and also with the MultiCTC approach described in Section~\ref{sec:multi_ctc}.
The results are summarized in Table~\ref{tab:results_comparison}.
Both filtering-free pseudo-labeling approaches (MultiCTC and SoftCTC) beat AT-ST without filtering (AT-ST 100\,\%) and reach the performance of the optimal filtering setting (AT-ST 10\,\%).
SoftCTC consistently achieves slightly better results than MultiCTC.

\begin{table}[t]
    \centering
    \caption{Final results comparison, reported as CER [\%] on the validation and test sets of the Bentham-Full dataset.
    The best result of the SoftCTC is selected based on the validation performance from the previous experiments.}
    \label{tab:results_comparison}
    \begin{tabular}{
        @{\extracolsep{4pt}}@{\kern\tabcolsep}
        lcc}
        \toprule
        
         & Val & Test \\
        \midrule
        Seed model  & 6.60 & 6.43 \\
        \midrule
        AT-ST 10\,\%  & 4.78 & 4.41 \\
        AT-ST 100\,\% & 4.96 & 4.73 \\
        \midrule
        MultiCTC & 4.80 & 4.40 \\
        SoftCTC & \textbf{4.48} & \textbf{4.35} \\
        \bottomrule
    \end{tabular}
\end{table}

\subsubsection*{Conclusion of experiments}
In this group of experiments, we have explored the properties and the robustness of the proposed SoftCTC loss.
Of all the tested enhancements, only smoothing probabilities brings a tangible improvement -- and still a modest one.
Based on these results, we conclude that SoftCTC is quite robust and utilizes the variability in the output of the CTC decoder well.
When compared to AT-ST, SoftCTC outperformed the AT-ST 100\,\% and reached the results of the AT-ST 10\,\% where a finely-tuned confidence-based filtration is used. 

\subsection{Generality and data efficiency}
\label{sec:experiments_group2}

We conducted self-training experiments with the SoftCTC loss also on the READ'16, Rodrigo, and Bentham-20k datasets.
Training of the optical model in each of these experiments consists of three stages: (1) seed model training, (2) fine-tuning, and (3) self-training.
In the seed model training stage, the model is trained similarly as in the previous experiments described in \ref{sec:experiments_group1}, except for the training data.
As the training data, we took different amounts of randomly selected lines (128, 256, 512, 1024, and 2048) from the training part of a given target dataset together with the related dataset READ'17.
We considered the rest of the lines in the training part of each target dataset as unannotated ones and we used them in the self-training stage.

In recent works~\cite{kohut_finetuning_2023, kohut_towards_2023}, a simple fine-tuning of a model on the target domain dataset showed great success even for a really small number of lines.
Therefore, we propose a fine-tuning stage, where the model is trained on the annotated lines from the target domain only.

Finally, during the self-training stage, we use the current model (at first the fine-tuned model, then the one from the previous self-training cycle) to obtain confusion networks for unannotated lines, and we continue training the current model on the mixture of the human-annotated and the machine-annotated lines from the target domain.
In total, we always perform three self-training cycles.

Additionally, as a topline, we trained a seed model on the READ'17 and all annotated training lines for each target dataset to obtain an upper-bound character error rate for comparison.

In the finetuning stage, we trained the model with a learning rate $5 \times 10^{-5}$ and batch size 32 for 10k iterations.
When fine-tuning the topline, we performed 20k iterations as the models needed more iterations to converge.
In the self-training stage, we trained the model for 25k iterations in each cycle with batch size 32.
In the first two cycles, we used the learning rate $1 \times 10^{-4}$, and in the last cycle, we used the learning rate $5 \times 10^{-5}$.

To decode the transcribed text lines, we use the partial line decoding strategy with a beam size of 16.
As shown in the previous experiments and also in preliminary experiments on the Rodrigo dataset, the difference between the two decoding strategies is rather insignificant, but the partial line strategy is considerably faster.

\begin{figure*}
    \centering
    
    \includegraphics{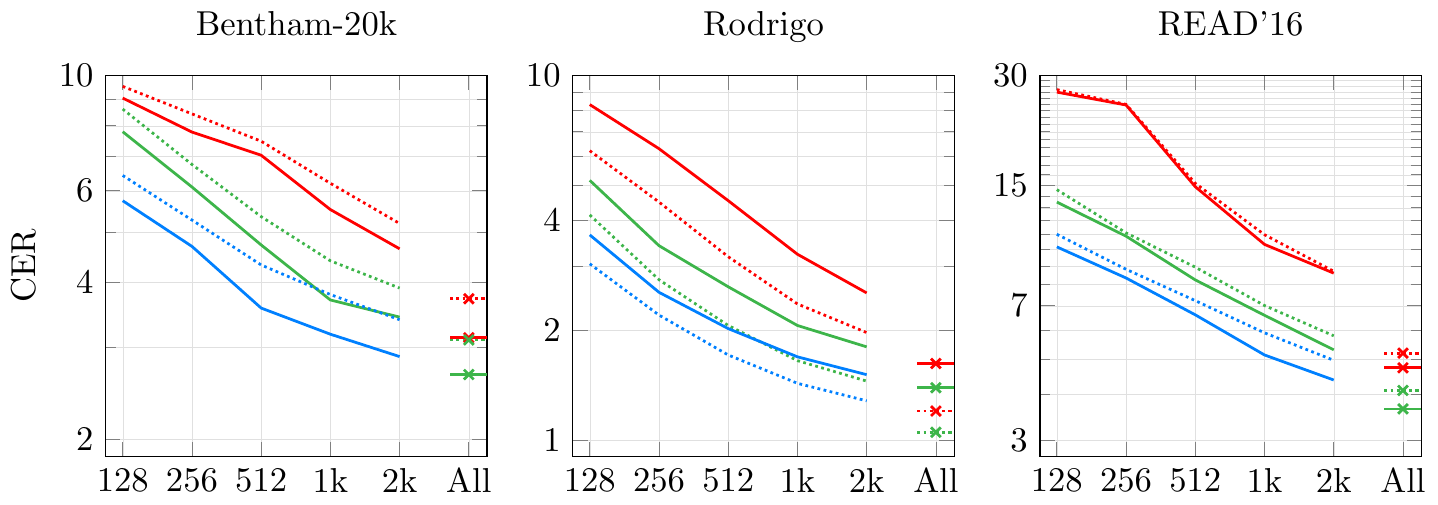}
    \\
    Number of annotated training lines 
    \vspace{0.1cm} \\
    \includegraphics{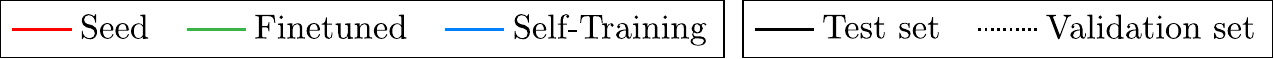}    
    \vspace{0.1cm} \\
    
    \caption{
    Results of self-training experiments on the Bentham-20k, Rodrigo, and READ'16 datasets.
    The horizontal axis represents a different amount of annotated training lines from each target dataset and the vertical axis represents CER in \% (note the different scales on all three vertical axes).
    The solid and dotted lines show the results on the test and validation sets, respectively.
    The results of the seed models are in red, the results of fine-tuned models are in green, and the results of models after self-training are in blue.
    The red and green crosses accompanied by short horizontal lines in the \emph{All} columns represent the results of the seed and fine-tuned models trained in a fully supervised manner on all annotated training lines (i.e. the upper bound).
    }
    \label{fig:finetuning_and_self-training_experiments}
\end{figure*}

Results of the experiments are shown in Fig.~\ref{fig:finetuning_and_self-training_experiments}.
The self-training on the unannotated lines with SoftCTC consistently improves the fine-tuned models.
Moreover, on the Rodrigo and Bentham-20k test sets, the self-training almost closes the gap between the topline -- fine-tuned on all of the training lines -- and the self-trained model with 2048 initial training lines.

\subsection{SoftCTC computational efficiency}
\label{sec:experiments_performance-experiments}

Besides the demonstration of the effectiveness of the proposed SoftCTC loss in training a neural network in the pseudo-labeling approach, we also show a comparison with the CTC loss (more precisely, the MultiCTC approach described in Section~\ref{sec:multi_ctc}) in terms of computational efficiency.
We compare CUDA implementations of PyTorch's CTC loss\footnote{\texttt{torch.nn.CTCLoss}} and the SoftCTC loss on three GPUs: NVIDIA GeForce 1080 Ti, NVIDIA GeForce RTX 2080 Ti, and NVIDIA GeForce RTX 3080.
We measure the time of both loss functions for batch sizes of powers of two from 4 to 256.
While we calculate the CTC loss w.r.t. only a single transcription, the SoftCTC is evaluated on confusion networks generated using the full line decoding strategy with beam size 16.
As we mention in Section~\ref{sec:multi_ctc}, calculating MultiCTC (i.e. calculating CTC w.r.t. all transcription variants) would require either more time or more memory, both proportional to the number of transcription variants.
In the first case, where the transcription variants are handled sequentially, it is needed to multiply the time of the CTC loss by the beam size $B$, which is the lower bound on the number of transcriptions represented in the appropriate confusion network and at the same time, it is an upper bound on the number of alternatives in each confusion set.
In practice, the amount of transcriptions in a confusion network is almost always much higher than the lower bound.
In the second case, the CTC loss can be calculated w.r.t. all transcription variants in parallel (i.e. the transcription variants are treated as independent text lines), but it would affect the memory usage and thus it can reach the hardware limits quickly.

\begin{figure*}
    \centering
    \includegraphics{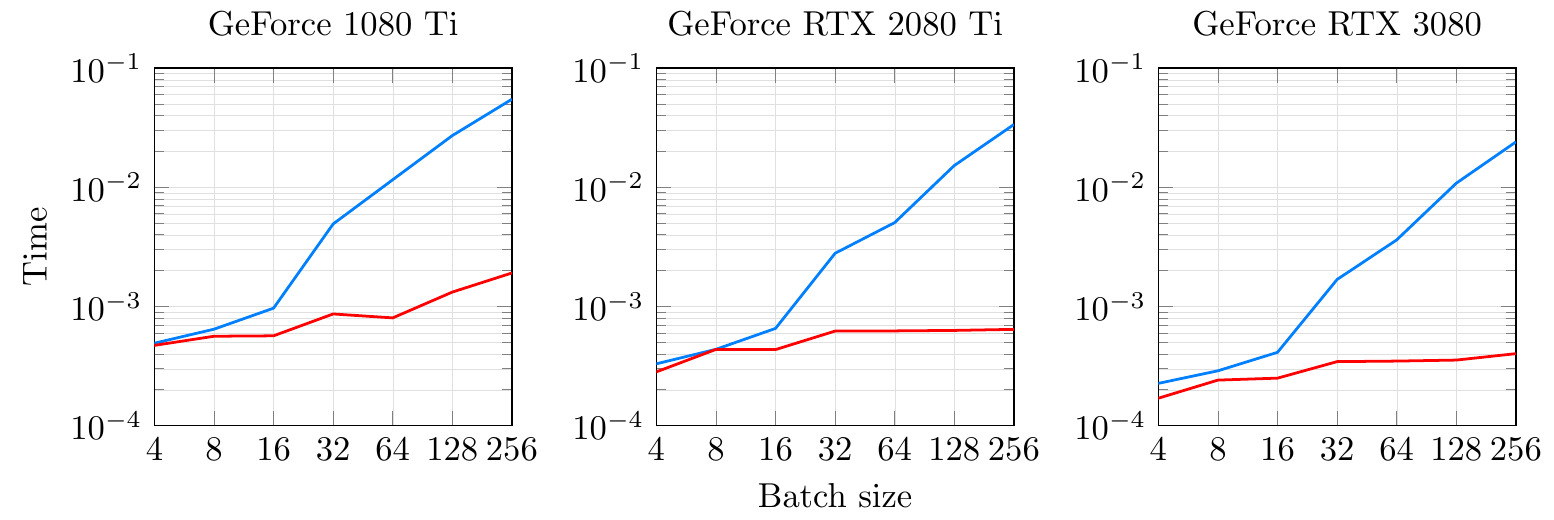}
    \vspace{0.1cm}
    \includegraphics{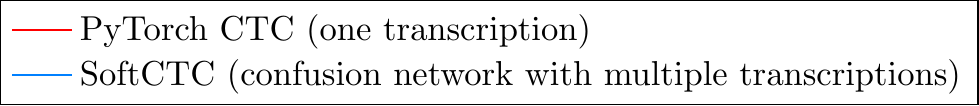}
    \caption{
        Comparison of the computational efficiency of SoftCTC and CTC (as implemented in PyTorch).
        Time in seconds per single batch of a given size.
        SoftCTC is operating on confusion networks generated from prefix search decoding with beam size 16, so it is computing the loss w.r.t. many transcriptions compared to the single transcription considered by CTC.
    }
    \label{fig:performance}
\end{figure*}

The measured results of both loss functions are depicted in Fig.~\ref{fig:performance}.
Absolute values for SoftCTC and MultiCTC are summarized in Table~\ref{tab:performance_comparison} for practical batch sizes and the most powerful GPU.
SoftCTC is significantly faster, esp. for smaller batches that need to be used with large models.
As mentioned before, we assume the lower bound on the number of transcription variants considered when estimating the MultiCTC computation time.

\begin{table}[t]
    \centering
    \caption{Time efficiency comparison [ms] on RTX 3080 with beam size 16.
        Values for CTC and SoftCTC are directly taken from Fig.~\ref{fig:performance} (GeForce RTX 3080), values for MultiCTC are obtained by multiplying the CTC values by the beam size.
    }
    \label{tab:performance_comparison}
    \begin{tabular}{
        @{\extracolsep{4pt}}@{\kern\tabcolsep}
        lccc}
        \toprule
        & \multicolumn{3}{c}{Batch size} \\
            \cline{2-4}
         & 16 & 32 & 64 \\
        \midrule
        CTC      & 0.250 & 0.345 & 0.348 \\
        MultiCTC & 4.006 & 5.513 & 5.572 \\
        SoftCTC  & \textbf{0.412} & \textbf{1.685} & \textbf{3.621} \\
        \bottomrule
    \end{tabular}
\end{table}

\section{Conclusion}
\label{sec:conclusion}
We have formulated an extended version of the popular CTC loss -- SoftCTC -- which allows to consider multiple target labelings at the same time.
Using SoftCTC, we were able to propose a novel approach to semi-supervised training of sequence models, where no explicit filtering of pseudo-labels is needed.
This allows to avoid a complex step in the pseudo-labeling training pipeline.

On the task of handwriting recognition, we have experimentally verified that our filtration-free approach matches the performance of a finely-tuned filtration-based pseudo-labeling pipeline.
More specifically, we have shown that it is advantageous to obtain the transcription variants locally in low-confidence regions of the output rather than from running beam-search on whole text lines.
Additionally, we have found it is beneficial to smooth out the output probabilities before summarizing them in a confusion network.
Furthermore, we tried to increase the variability in the transcription variants by aggregating outputs from multiple augmented versions of the same text line.
However, this did not lead to improvements.
Finally, we have shown that no significant benefits can be achieved by filtering text lines in our pipeline, supporting the hypothesis that our soft pseudo-labels-based approach is robust in and of itself.

Furthermore, we have demonstrated the effectiveness of SoftCTC on a triplet of datasets, exploring how well it scales with increasing amounts of transcribed data.
SoftCTC improves quickly with additional human-annotation, in two cases virtually reaching the topline performance when 2048 transcribed lines are provided.

We have compared the computational cost of using SoftCTC to the cost of a naïve repeated CTC solution.
We found out that SoftCTC implementation is superior for a wide range of conditions.
We make our SoftCTC implementation public, including an efficient CUDA version.

We believe that SoftCTC will make semi-supervised training a more available method for practitioners in the field, allowing better domain-specific OCR systems to be deployed.
SoftCTC is not limited to OCR but can be applied to any sequence transduction task, such as ASR.
It remains a future work to extend the ideas of SoftCTC to RNN-T, which dominates the field of streaming ASR nowadays.

\section*{Declarations}

The authors do not have any competing interests, financial or other.

\bibliography{sn-bibliography}

\begin{thebibliography}{45}
\providecommand{\natexlab}[1]{#1}
\providecommand{\url}[1]{{#1}}
\providecommand{\urlprefix}{URL }
\providecommand{\doi}[1]{\url{https://doi.org/#1}}
\providecommand{\eprint}[2][]{\url{#2}}
 \bibcommenthead

\bibitem[{Shi et~al(2017)Shi, Bai, and Yao}]{shi_end--end_2017}
Shi B, Bai X, Yao C (2017) An end-to-end trainable neural network for
  image-based sequence recognition and its application to scene text
  recognition. {IEEE} Transactions on Pattern Analysis and Machine Intelligence
  39(11):2298--2304. Conference Name: {IEEE} Transactions on Pattern Analysis
  and Machine Intelligence

\bibitem[{Graves et~al(2006)Graves, Fern\'{a}ndez, Gomez, and
  Schmidhuber}]{GravesCTC}
Graves A, Fern\'{a}ndez S, Gomez F, et~al (2006) Connectionist {T}emporal
  {C}lassification: {L}abelling {U}nsegmented {S}equence {D}ata with
  {R}ecurrent {N}eural {N}etworks. In: Proceedings of the 23rd International
  Conference on Machine Learning, New York, NY, USA, ICML '06, p 369–376

\bibitem[{Sutskever et~al(2014)Sutskever, Vinyals, and Le}]{SutskeverSeq2Seq}
Sutskever I, Vinyals O, Le QV (2014) Sequence to sequence learning with neural
  networks. In: Ghahramani Z, Welling M, Cortes C, et~al (eds) Advances in
  Neural Information Processing Systems, vol~27. Curran Associates, Inc.

\bibitem[{Radford et~al(2022)Radford, Kim, Xu, Brockman, {McLeavey}, and
  Sutskever}]{radford_robust_2022}
Radford A, Kim JW, Xu T, et~al (2022) Robust speech recognition via large-scale
  weak supervision.

\bibitem[{Brown et~al(2020)Brown, Mann, Ryder, Subbiah, Kaplan, Dhariwal,
  Neelakantan, Shyam, Sastry, Askell, Agarwal, Herbert-Voss, Krueger, Henighan,
  Child, Ramesh, Ziegler, Wu, Winter, Hesse, Chen, Sigler, Litwin, Gray, Chess,
  Clark, Berner, {McCandlish}, Radford, Sutskever, and
  Amodei}]{brown_language_2020}
Brown T, Mann B, Ryder N, et~al (2020) Language models are few-shot learners.
  In: Advances in Neural Information Processing Systems, vol~33. Curran
  Associates, Inc., pp 1877--1901

\bibitem[{Rombach et~al(2022)Rombach, Blattmann, Lorenz, Esser, and
  Ommer}]{rombach_high-resolution_2022}
Rombach R, Blattmann A, Lorenz D, et~al (2022) High-resolution image synthesis
  with latent diffusion models. In: Proceedings of the IEEE/CVF Conference on
  Computer Vision and Pattern Recognition (CVPR), pp 10,684--10,695

\bibitem[{Ramesh et~al(2022)Ramesh, Dhariwal, Nichol, Chu, and
  Chen}]{ramesh_hierarchical_2022}
Ramesh A, Dhariwal P, Nichol A, et~al (2022) Hierarchical text-conditional
  image generation with {CLIP} latents. \eprint{2204.06125 [cs]}

\bibitem[{Kišš et~al(2021)Kišš, Beneš, and Hradiš}]{kiss_at-st_2021}
Kišš M, Beneš K, Hradiš M (2021) {AT}-{ST}: Self-training adaptation
  strategy for {OCR} in domains with limited transcriptions. In: Lladós J,
  Lopresti D, Uchida S (eds) Document Analysis and Recognition – {ICDAR}
  2021. Springer International Publishing, pp 463--477

\bibitem[{Arazo et~al(2020)Arazo, Ortego, Albert, O’Connor, and
  {McGuinness}}]{arazo_pseudo-labeling_2020}
Arazo E, Ortego D, Albert P, et~al (2020) Pseudo-labeling and confirmation bias
  in deep semi-supervised learning. In: 2020 International Joint Conference on
  Neural Networks ({IJCNN}), pp 1--8, {ISSN}: 2161-4407

\bibitem[{Bachman et~al(2014)Bachman, Alsharif, and
  Precup}]{bachman_learning_2014}
Bachman P, Alsharif O, Precup D (2014) Learning with pseudo-ensembles.
  \eprint{1412.4864 [cs, stat]}

\bibitem[{Sajjadi et~al(2016)Sajjadi, Javanmardi, and
  Tasdizen}]{sajjadi_regularization_2016}
Sajjadi M, Javanmardi M, Tasdizen T (2016) Regularization with stochastic
  transformations and perturbations for deep semi-supervised learning. In:
  Advances in Neural Information Processing Systems, vol~29. Curran Associates,
  Inc.

\bibitem[{Tarvainen and Valpola(2017)}]{tarvainen_mean_2017}
Tarvainen A, Valpola H (2017) Mean teachers are better role models:
  Weight-averaged consistency targets improve semi-supervised deep learning
  results. In: Advances in Neural Information Processing Systems, vol~30.
  Curran Associates, Inc.

\bibitem[{Berthelot et~al(2019)Berthelot, Carlini, Goodfellow, Papernot,
  Oliver, and Raffel}]{berthelot_mixmatch_2019}
Berthelot D, Carlini N, Goodfellow I, et~al (2019) {MixMatch}: A holistic
  approach to semi-supervised learning. In: Advances in Neural Information
  Processing Systems, vol~32. Curran Associates, Inc.

\bibitem[{Kurakin et~al(2020)Kurakin, Raffel, Berthelot, Cubuk, Zhang, Sohn,
  and Carlini}]{kurakin_remixmatch_2020}
Kurakin A, Raffel C, Berthelot D, et~al (2020) {ReMixMatch}: Semi-supervised
  learning with distribution matching and augmentation anchoring. In: {ICLR}

\bibitem[{Xie et~al(2020)Xie, Dai, Hovy, Luong, and Le}]{xie_unsupervised_2020}
Xie Q, Dai Z, Hovy E, et~al (2020) Unsupervised data augmentation for
  consistency training. In: Larochelle H, Ranzato M, Hadsell R, et~al (eds)
  Advances in Neural Information Processing Systems, vol~33. Curran Associates,
  Inc., pp 6256--6268

\bibitem[{Englesson and Azizpour(2021)}]{englesson_generalized_2021}
Englesson E, Azizpour H (2021) Generalized jensen-shannon divergence loss for
  learning with noisy labels. In: Advances in Neural Information Processing
  Systems, vol~34. Curran Associates, Inc., pp 30,284--30,297

\bibitem[{Zheng et~al(2022)Zheng, Li, Rhee, Han, Han, and
  Wang}]{zheng_pushing_2022}
Zheng C, Li H, Rhee S, et~al (2022) Pushing the performance limit of scene text
  recognizer without human annotation. In: 2022 {IEEE}/{CVF} Conference on
  Computer Vision and Pattern Recognition ({CVPR}), pp 14,096--14,105, {ISSN}:
  2575-7075

\bibitem[{Aberdam et~al(2022)Aberdam, Ganz, Mazor, and
  Litman}]{aberdam_multimodal_2022}
Aberdam A, Ganz R, Mazor S, et~al (2022) Multimodal semi-supervised learning
  for text recognition. \eprint{2205.03873 [cs]}

\bibitem[{Lee(2013)}]{lee_pseudo-label_2013}
Lee DH (2013) Pseudo-label : The simple and efficient semi-supervised learning
  method for deep neural networks. {ICML} 2013 Workshop : Challenges in
  Representation Learning ({WREPL})

\bibitem[{Xie et~al(2020)Xie, Luong, Hovy, and Le}]{xie_self-training_2020}
Xie Q, Luong MT, Hovy E, et~al (2020) Self-{Training} {With} {Noisy} {Student}
  {Improves} {ImageNet} {Classification}. In: Proceedings of the IEEE/CVF
  Conference on Computer Vision and Pattern Recognition (CVPR)

\bibitem[{Pham et~al(2021)Pham, Dai, Xie, and Le}]{pham_meta_2021}
Pham H, Dai Z, Xie Q, et~al (2021) Meta pseudo labels. In: Proceedings of the
  IEEE/CVF Conference on Computer Vision and Pattern Recognition (CVPR), pp
  11,557--11,568

\bibitem[{Nagai(2020)}]{nagai_recognizing_2020}
Nagai A (2020) Recognizing {Japanese} {Historical} {Cursive} with
  {Pseudo}-{Labeling}-aided {CRNN} as an {Application} of {Semi}-{Supervised}
  {Learning} to {Sequence} {Labeling}. In: 2020 17th {International}
  {Conference} on {Frontiers} in {Handwriting} {Recognition} ({ICFHR}), pp
  97--102

\bibitem[{Stuner et~al(2017)Stuner, Chatelain, and
  Paquet}]{stuner_self-training_2017}
Stuner B, Chatelain C, Paquet T (2017) Self-{Training} of {BLSTM} with
  {Lexicon} {Verification} for {Handwriting} {Recognition}. In: 2017 14th
  {IAPR} {International} {Conference} on {Document} {Analysis} and
  {Recognition} ({ICDAR}), pp 633--638, iSSN: 2379-2140

\bibitem[{Leifert et~al(2020)Leifert, Labahn, and Sánchez}]{leifert_two_2020}
Leifert G, Labahn R, Sánchez JA (2020) Two {Semi}-{Supervised} {Training}
  {Approaches} for {Automated} {Text} {Recognition}. In: 2020 17th
  {International} {Conference} on {Frontiers} in {Handwriting} {Recognition}
  ({ICFHR}), pp 145--150

\bibitem[{Das and Jawahar(2020)}]{das_adapting_2020}
Das D, Jawahar CV (2020) Adapting {OCR} with {Limited} {Supervision}. In: Bai
  X, Karatzas D, Lopresti D (eds) Document {Analysis} {Systems}. Springer
  International Publishing, Cham, Lecture {Notes} in {Computer} {Science}, pp
  30--44

\bibitem[{Sohn et~al(2020)Sohn, Berthelot, Carlini, Zhang, Zhang, Raffel,
  Cubuk, Kurakin, and Li}]{sohn_fixmatch_2020}
Sohn K, Berthelot D, Carlini N, et~al (2020) {FixMatch}: Simplifying
  semi-supervised learning with consistency and confidence. In: Advances in
  Neural Information Processing Systems, vol~33. Curran Associates, Inc., pp
  596--608

\bibitem[{Weninger et~al(2020)Weninger, Mana, Gemello, Andrés-Ferrer, and
  Zhan}]{weninger_semi-supervised_2020}
Weninger F, Mana F, Gemello R, et~al (2020) Semi-supervised learning with data
  augmentation for end-to-end {ASR}. In: Proc. Interspeech 2020, pp 2802--2806

\bibitem[{Wolf and Fink(2022)}]{wolf_self-training_2022}
Wolf F, Fink GA (2022) Self-training of handwritten word recognition for
  synthetic-to-real adaptation. In: Proc. Int. Conf. on Pattern Recognition

\bibitem[{Zhang et~al(2018)Zhang, Cisse, Dauphin, and
  Lopez-Paz}]{zhang_mixup_2018}
Zhang H, Cisse M, Dauphin YN, et~al (2018) mixup: Beyond empirical risk
  minimization. In: International Conference on Learning Representations

\bibitem[{Frinken and Bunke(2009)}]{frinken_evaluating_2009}
Frinken V, Bunke H (2009) Evaluating retraining rules for semi-supervised
  learning in neural network based cursive word recognition. In: 2009 10th
  International Conference on Document Analysis and Recognition, pp 31--35,
  {ISSN}: 2379-2140

\bibitem[{Constum et~al(2022)Constum, Kempf, Paquet, Tranouez, Chatelain,
  Brée, and Merveille}]{constum_recognition_2022}
Constum T, Kempf N, Paquet T, et~al (2022) Recognition and {I}nformation
  {E}xtraction in {H}istorical {H}andwritten {T}ables: {T}oward
  {U}nderstanding {E}arly $20^{\mathrm{th}}$ {C}entury {P}aris {C}ensus. In:
  Uchida S, Barney E, Eglin V (eds) Document Analysis Systems. Springer
  International Publishing, Lecture Notes in Computer Science, pp 143--157

\bibitem[{Gao et~al(2021)Gao, Chen, Wang, and Lu}]{gao_semi-supervised_2021}
Gao Y, Chen Y, Wang J, et~al (2021) Semi-supervised scene text recognition.
  {IEEE} Transactions on Image Processing 30:3005--3016. Conference Name:
  {IEEE} Transactions on Image Processing

\bibitem[{Soltau et~al(2017)Soltau, Liao, and Sak}]{Soltau-CTC-ASR-2017}
Soltau H, Liao H, Sak H (2017) Neural speech recognizer: Acoustic-to-word lstm
  model for large vocabulary speech recognition. In: Proc. Interspeech 2017, pp
  3707--3711

\bibitem[{Watanabe et~al(2017)Watanabe, Hori, Kim, Hershey, and
  Hayashi}]{WatanabeHybridCTC}
Watanabe S, Hori T, Kim S, et~al (2017) Hybrid ctc/attention architecture for
  end-to-end speech recognition. IEEE Journal of Selected Topics in Signal
  Processing 11(8):1240--1253

\bibitem[{Rabiner(1989)}]{Rabiner1989}
Rabiner LR (1989) A tutorial on hidden {M}arkov models and selected
  applications in speech recognition. Proceedings of the IEEE 77(2):257--286

\bibitem[{Bangalore et~al(2001)Bangalore, Bordel, and
  Riccardi}]{bangalore_computing_2001}
Bangalore B, Bordel G, Riccardi G (2001) Computing consensus translation from
  multiple machine translation systems. In: {IEEE} Workshop on Automatic Speech
  Recognition and Understanding, 2001. {ASRU} '01., pp 351--354

\bibitem[{Rosti et~al(2007)Rosti, Ayan, Xiang, Matsoukas, Schwartz, and
  Dorr}]{rosti_combining_2007}
Rosti AV, Ayan NF, Xiang B, et~al (2007) Combining outputs from multiple
  machine translation systems. In: Human Language Technologies 2007: The
  Conference of the North American Chapter of the Association for Computational
  Linguistics; Proceedings of the Main Conference. Association for
  Computational Linguistics, pp 228--235

\bibitem[{Fiscus(1997)}]{fiscus_post-processing_1997}
Fiscus J (1997) A post-processing system to yield reduced word error rates:
  Recognizer output voting error reduction ({ROVER}). In: 1997 {IEEE} Workshop
  on Automatic Speech Recognition and Understanding Proceedings, pp 347--354

\bibitem[{Mangu et~al(2000)Mangu, Brill, and Stolcke}]{mangu_finding_2000}
Mangu L, Brill E, Stolcke A (2000) Finding consensus in speech recognition:
  word error minimization and other applications of confusion networks.
  Computer Speech \& Language 14(4):373--400

\bibitem[{Sanchez et~al(2017)Sanchez, Romero, Toselli, Villegas, and
  Vidal}]{sanchez_icdar2017_2017}
Sanchez JA, Romero V, Toselli AH, et~al (2017) {ICDAR2017} {Competition} on
  {Handwritten} {Text} {Recognition} on the {READ} {Dataset}. In: 2017 14th
  {IAPR} {International} {Conference} on {Document} {Analysis} and
  {Recognition} ({ICDAR}). IEEE, Kyoto, pp 1383--1388

\bibitem[{Sánchez et~al(2014)Sánchez, Romero, Toselli, and
  Vidal}]{sanchez_icfhr2014_2014}
Sánchez JA, Romero V, Toselli AH, et~al (2014) {ICFHR2014} {Competition} on
  {Handwritten} {Text} {Recognition} on {Transcriptorium} {Datasets} ({HTRtS}).
  In: 2014 14th {International} {Conference} on {Frontiers} in {Handwriting}
  {Recognition}, pp 785--790, iSSN: 2167-6445

\bibitem[{Serrano et~al(2010)Serrano, Castro, and Juan}]{serrano_rodrigo_2010}
Serrano N, Castro F, Juan A (2010) The {RODRIGO} database. In: Proceedings of
  the Seventh International Conference on Language Resources and Evaluation
  ({LREC}'10). European Language Resources Association ({ELRA})

\bibitem[{Sánchez et~al(2016)Sánchez, Romero, Toselli, and
  Vidal}]{sanchez_icfhr2016_2016}
Sánchez JA, Romero V, Toselli AH, et~al (2016) {ICFHR}2016 competition on
  handwritten text recognition on the {READ} dataset. In: 2016 15th
  International Conference on Frontiers in Handwriting Recognition ({ICFHR}),
  pp 630--635, {ISSN}: 2167-6445

\bibitem[{Kohút and Hradiš(2023)}]{kohut_finetuning_2023}
Kohút J, Hradiš M (2023) Finetuning is a surprisingly effective domain
  adaptation baseline in handwriting recognition. \eprint{2302.06308 [cs]}

\bibitem[{Kohút et~al(2023)Kohút, Hradiš, and Kišš}]{kohut_towards_2023}
Kohút J, Hradiš M, Kišš M (2023) Towards writing style adaptation in
  handwriting recognition. \eprint{2302.06318 [cs]}

\end{thebibliography}


\end{document}